\documentclass[11pt]{article}
\usepackage[preprint]{acl}

\usepackage{times}
\usepackage{latexsym}
\usepackage[T1]{fontenc}
\usepackage[utf8]{inputenc}
\pdfmapline{+optimistic < assets/Optimistic.ttf <T1-WGL4.enc}
\DeclareFontFamily{T1}{optimistic}{}
\DeclareFontShape{T1}{optimistic}{m}{n}{<-> s * [0.88] assets/optimistic}{}
\DeclareFontShape{T1}{optimistic}{b}{n}{<-> s * [0.88] assets/optimistic}{}

\usepackage{microtype}
\DisableLigatures[f]{family=sf*}
\usepackage{inconsolata}
\usepackage{graphicx}
\usepackage{booktabs}
\usepackage{colortbl}
\usepackage{multirow}
\usepackage{amsmath,amssymb,amsthm}
\usepackage{enumitem}
\usepackage{algorithm}
\usepackage{algpseudocode}
\usepackage{xspace}
\usepackage{placeins}
\usepackage{etoolbox}
\usepackage{tabularx}
\usepackage[most]{tcolorbox}
\usepackage{fontawesome5}
\definecolor{EVGroup}{RGB}{240,238,245}
\definecolor{EVRow}{RGB}{230,241,248}
\definecolor{EVCell}{RGB}{209,230,243}
\definecolor{ArxivCardBG}{HTML}{F5F8FF}
\definecolor{ArxivInk}{RGB}{31,29,39}
\definecolor{ArxivBlue}{RGB}{35,74,210}
\definecolor{ArxivLink}{HTML}{0F297F}
\definecolor{ArxivOrange}{RGB}{255,122,0}
\hypersetup{
  pdftitle={Length-Adaptive Decoding for Masked Diffusion Machine Translation},
  pdfauthor={Yan Zhan, Mengkai Hou, Wanting Zhang, Zhijun Gao},
  citecolor=ArxivLink,
  linkcolor=ArxivLink,
  urlcolor=ArxivLink
}
\newcommand{\V}{\mathcal{V}}
\newcommand{\Mask}{[\textsc{Mask}]}
\newcommand{\xvec}{\mathbf{x}}
\newcommand{\yvec}{\mathbf{y}}
\newcommand{\Tsteps}{T}
\newcommand{\Lstar}{L^{\star}}
\newcommand{\Lref}{L^{*}}
\newcommand{\Lhat}{\widehat{L}}
\newcommand{\Cset}{\mathcal{C}}
\newcommand{\Rset}{\mathcal{R}}
\newcommand{\Hent}{H}
\newcommand{\Hbar}{\bar{H}}
\DeclareMathOperator*{\argmin}{arg\,min}

\newcommand{\oracleref}{Length Oracle\xspace}
\newcommand{\ratio}{Ratio\xspace}
\newcommand{\method}{Entropy-Valley\xspace}

\newcommand{\arxivabstract}{%
Machine translation tests masked diffusion language models (dLLMs) because every source token must be rendered faithfully, while fixed canvas decoding must choose target length before denoising.
Existing masked diffusion decoding work mainly studies token unmasking order, leaving this length decision under-explored despite its direct effect on coverage and redundancy.
We introduce \method (EV), a training-free length selector that scores candidate target canvases by mean predictive entropy from all-mask forward passes and selects the canvas the backbone is most prepared to fill.
Relative to a baseline using training corpus length statistics, EV recovers $64.9\%$, $65.3\%$, and $33.0\%$ of the COMET-22 gain from reference target lengths on \mbox{En$\to$Zh}, \mbox{Zh$\to$En}, and \mbox{En$\to$De}. Our diagnostics show that denoising-friendly lengths need not match reference lengths.
Evaluation by three translation experts supports the \mbox{En$\leftrightarrow$Zh} adequacy gains, with stronger evidence on \mbox{Zh$\to$En}. Compared with a LLaMA-3-8B autoregressive (AR) model trained on the same fine-tuning data, the EV system ties on \mbox{En$\to$Zh} and leads on \mbox{Zh$\to$En}; an oracle-length diagnostic further shows that, in this masked diffusion MT setting, deciding which tokens to reveal first matters less than how the target length is supplied.%
}

\newcommand{\sfauthmark}[1]{\raisebox{0.55ex}{\fontsize{6}{6}\selectfont\sffamily #1}}
\newcommand{\rmaffmark}[1]{\raisebox{0.55ex}{\fontsize{6}{6}\selectfont\rmfamily #1}}
\newcommand{\metadataicon}[2]{%
  \makebox[1.35em][c]{\raisebox{-0.10em}{\textcolor{#1}{#2}}}%
}
\newcommand{\hfmetadataicon}{%
  \makebox[1.35em][c]{\raisebox{-0.23em}{\includegraphics[height=1.16em]{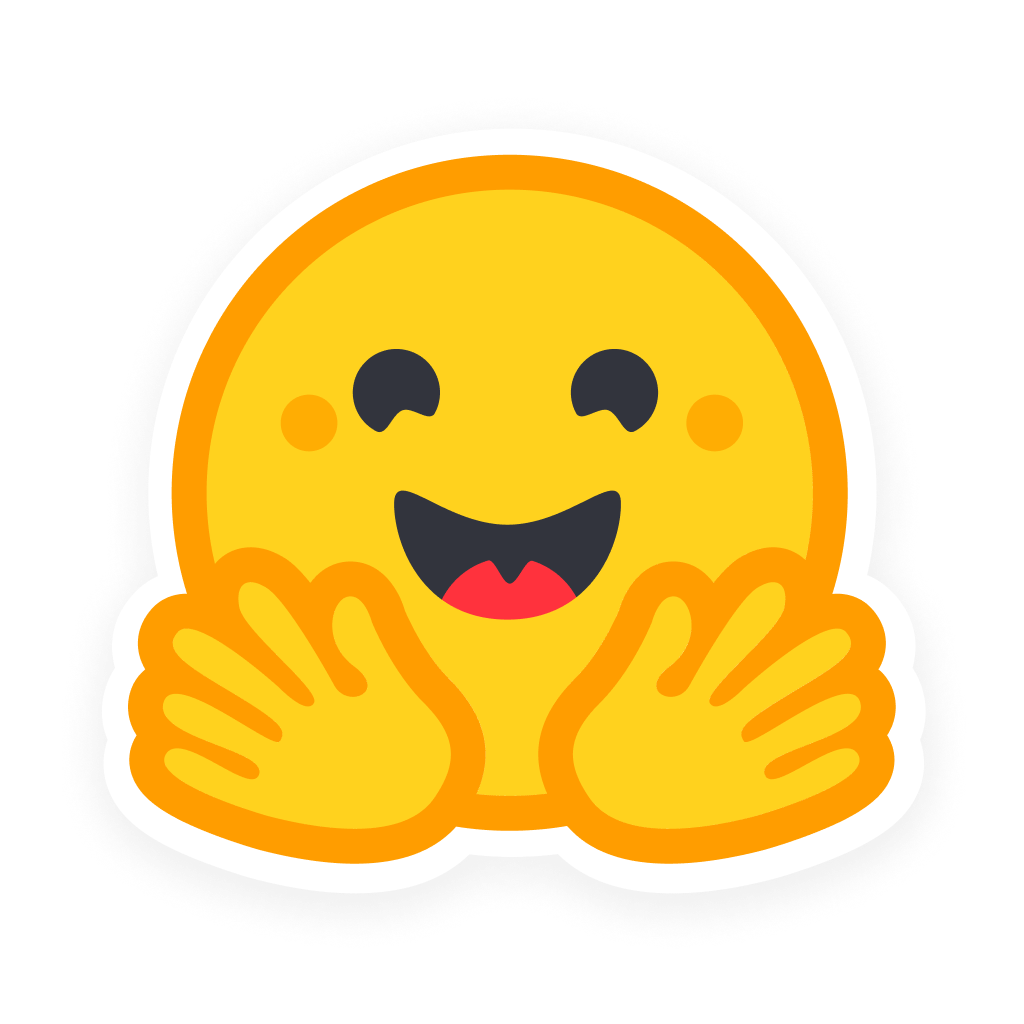}}}%
}

\DeclareTextFontCommand{\textbf}{\bfseries\sffamily}
\DeclareCaptionLabelSeparator{arxivcustom}{}
\DeclareCaptionFormat{arxivcustom}{{\sffamily\bfseries #1 #2} #3}
\DeclareCaptionFont{arxivref}{\fontsize{9}{11}\selectfont}
\floatname{algorithm}{\sffamily Algorithm}
\AtBeginEnvironment{thebibliography}{\fontsize{9}{11}\selectfont}
\makeatletter
\renewcommand\floatc@ruled[2]{%
  \fontsize{9}{11}\selectfont
  {\sffamily\bfseries #1} #2\par}
\def\section{\@startsection {section}{1}{\z@}{-1.9ex plus
    -0.5ex minus -.2ex}{1.4ex plus 0.3ex minus .2ex}{\Large\sffamily\bfseries\raggedright}}
\def\subsection{\@startsection{subsection}{2}{\z@}{-1.7ex plus
    -0.5ex minus -.2ex}{0.7ex plus .2ex}{\large\sffamily\bfseries\raggedright}}
\def\subsubsection{\@startsection{subsubsection}{3}{\z@}{-1.4ex plus
   -0.5ex minus -.2ex}{0.4ex plus .2ex}{\normalsize\sffamily\bfseries\raggedright}}
\def\paragraph{\@startsection{paragraph}{4}{\z@}{1.3ex plus
   0.5ex minus .2ex}{-1em}{\normalsize\sffamily\bfseries}}
\def\subparagraph{\@startsection{subparagraph}{5}{\parindent}{1.3ex plus
   0.5ex minus .2ex}{-1em}{\normalsize\sffamily\bfseries}}
\makeatother

\newcommand{\makearxivtitle}{%
\twocolumn[{%
\begin{tcolorbox}[
  enhanced,
  width=\textwidth,
  colback=ArxivCardBG,
  frame hidden,
  arc=10pt,
  outer arc=10pt,
  boxsep=0pt,
  left=5mm,
  right=5mm,
  top=5mm,
  bottom=5mm,
  overlay={%
    \node[anchor=south east,inner sep=0pt]
      at ([xshift=-5mm,yshift=6.5mm]frame.south east) {%
        \includegraphics[height=1.00cm]{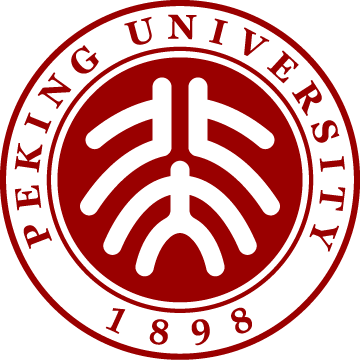}%
        \hspace{2.1mm}%
        \includegraphics[height=1.00cm]{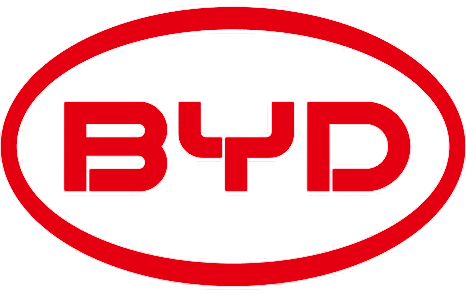}%
      };%
  }
]
\color{ArxivInk}
\noindent
\begin{minipage}[c][2.15cm][c]{0.20\linewidth}
  \centering
  \includegraphics[width=2.95cm]{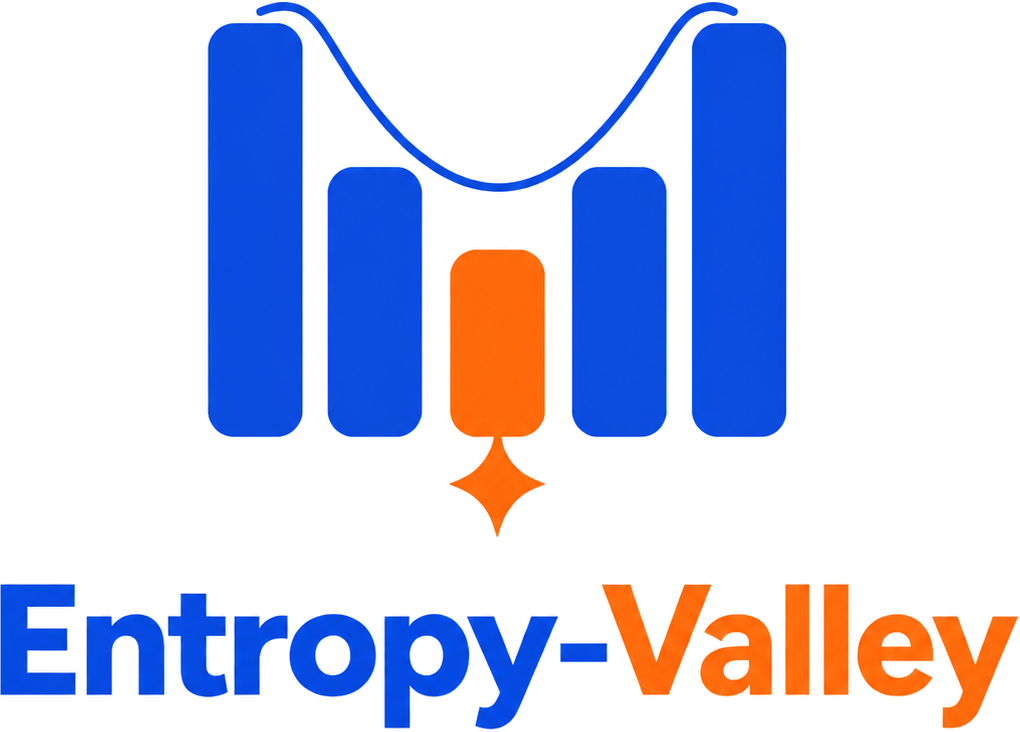}
\end{minipage}\hfill
\begin{minipage}[c][2.15cm][c]{0.785\linewidth}
  {\fontsize{20.2}{21.5}\selectfont\sffamily\bfseries
   \textcolor{ArxivBlue}{Length-Adaptive Decoding} for\\[-0.10ex]
   \mbox{Masked Diffusion Machine Translation}\par}
  \vspace{1.4mm}
  {\fontsize{10}{12}\selectfont\sffamily\bfseries
   Yan Zhan\sfauthmark{1}\enspace
   Mengkai Hou\sfauthmark{1}\enspace
   Wanting Zhang\sfauthmark{2}\enspace
   Zhijun Gao\sfauthmark{1}\par}
  \vspace{0.4mm}
  {\fontsize{8.8}{11}\selectfont\rmfamily
   \rmaffmark{1}Peking University\qquad
   \rmaffmark{2}BYD Company Limited\par}
\end{minipage}

\vspace{3.8mm}
{\fontsize{10}{12}\selectfont\rmfamily
 \setlength{\parindent}{0pt}
 \begingroup
 \renewcommand{\method}{\textbf{Entropy-Valley}\xspace}%
 \arxivabstract\par
 \endgroup}

\vspace{4.0mm}
\begin{minipage}[b]{0.80\linewidth}
  \fontsize{9}{11}\selectfont
  \urlstyle{same}
  \setlength{\tabcolsep}{0pt}
  \renewcommand{\arraystretch}{1.12}
  \begin{tabularx}{\linewidth}{@{}>{\centering\arraybackslash}p{0.45cm}@{}>{\sffamily\bfseries}p{2.75cm}@{\hspace{0.16cm}}X@{}}
    \metadataicon{ArxivInk}{\faEnvelope} & Contact: & \href{mailto:gaozhijun@pku.edu.cn}{gaozhijun@pku.edu.cn} \\
    \metadataicon{ArxivInk}{\faGithub} & Code: & \href{https://github.com/Entropy-Valley/Entropy-Valley}{\nolinkurl{github.com/Entropy-Valley/Entropy-Valley}} \\
    \hfmetadataicon & Dataset \& Model: & \href{https://huggingface.co/collections/YanZhanPKU/entropy-valley}{\nolinkurl{huggingface.co/collections/YanZhanPKU/entropy-valley}}
  \end{tabularx}
\end{minipage}
\end{tcolorbox}
\vspace{0.55\baselineskip}
}]%
}

\title{Length-Adaptive Decoding for Masked Diffusion Machine Translation}
\author{
Yan Zhan\textsuperscript{1}, Mengkai Hou\textsuperscript{1}, Wanting Zhang\textsuperscript{2}, Zhijun Gao\textsuperscript{1} \\
\textsuperscript{1}Peking University \quad
\textsuperscript{2}BYD Company Limited \\
\texttt{gaozhijun@pku.edu.cn}
}
\begin{document}
\makearxivtitle

\section{Introduction}
\label{sec:intro}

Pretrained masked diffusion language models (dLLMs) such as LLaDA-8B \citep{nie2025llada} and Dream-7B \citep{ye2025dream} now rival autoregressive (AR) LLMs on open ended generation and reasoning. Machine translation stresses a different part of the model. A translation system must preserve source content, reorder words across languages, choose suitable morphology, and stop at the right length. If the output is too short, content disappears; if it is too long, the model has room to repeat, hallucinate, or dilute the sentence.

The stopping problem is structural in fixed canvas masked diffusion decoding. An AR decoder generates tokens until it emits EOS. A masked diffusion decoder instead receives a target canvas before denoising begins. During supervised training, EOS is placed after that canvas, so at test time the model tends to fill whatever length it is given. Work on masked iterative generation and masked diffusion decoding often asks which positions to reveal first \citep{chang2022maskgit,hong2026oemdm,aman2026logicdiff}. Target length is often supplied either by the reference translation length during evaluation or by a single corpus-level source-to-target ratio. The former is an oracle upper bound rather than an inference procedure; the latter imposes one global length rule on sentences with different compression or expansion needs.

Borrowing length machinery from non autoregressive MT is not a clean fit for this setting. Classic systems predict fertilities or target lengths, sometimes with a small length beam \citep{gu2017nonautoregressive,ghazvininejad2019mask}, or change the decoder into an insertion and deletion architecture \citep{gu2019levenshtein}. Those options are reasonable when the model is trained for that purpose, but they do not answer a narrower question: whether a pretrained decoder only masked diffusion backbone can select its canvas at test time without another model or another training objective. Static corpus ratios have the opposite problem. They are easy to deploy, but one ratio cannot account for source sentences with placeholders, numbers, short commands, and idiomatic compression.

\begin{figure}[t]
  \centering
  \includegraphics[width=\columnwidth]{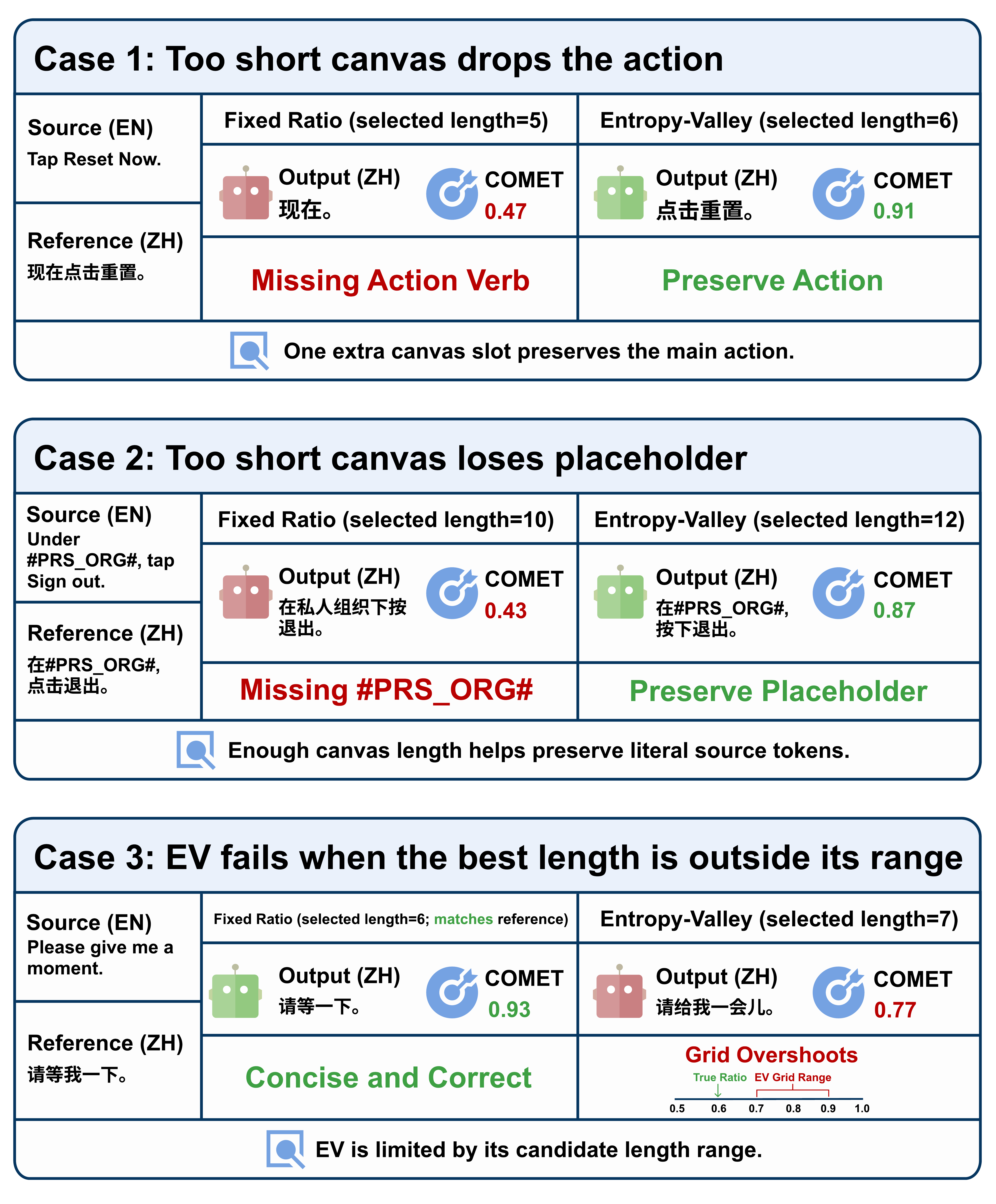}
  \caption{\textbf{Three \mbox{En$\to$Zh} cases decoded with 32 denoising steps.} A short canvas can drop an action verb or a required placeholder; EV chooses a longer canvas in these two cases and recovers the missing content. The third case shows the opposite boundary: the best compression lies below the EV candidate range, so the ratio baseline is better.}
  \label{fig:cases}
\end{figure}

Figure~\ref{fig:cases} shows why this is not a bookkeeping detail. For the source sentence ``Tap Reset Now.'', the ratio baseline chooses five target slots and drops the action verb; EV chooses six slots and recovers the verb. For ``Under \#PRS\_ORG\#, tap Sign out.'', the shorter canvas loses the placeholder, while the EV canvas preserves it. These are ordinary translation failures: the reader loses an action, an entity marker, or both. They arise before any question about the order in which masks are revealed.

The same pattern appears beyond these examples. On WMT22 \mbox{En$\to$Zh}, EV raises placeholder retention from $66.9\%$ to $89.1\%$ and number retention from $77.6\%$ to $81.3\%$ over the ratio baseline (Appendix~\ref{sec:appendix_coverage}). The gain is not just longer output. The right length depends on the source sentence: some inputs need extra room to preserve a placeholder, while high compression cases need a shorter canvas than our current candidate range provides. This points to length selection, not length inflation.

The model already exposes a useful signal before any token is generated. On an all-mask canvas, each candidate length induces a predictive distribution over every target slot. A canvas that is too short must compress too much source content into too few target slots; a canvas that is too long creates extra positions for which the model has no confident token choice. Mean predictive entropy provides a direct way to compare candidate lengths before spending the full decoding budget. Because this signal comes from the translation model itself, it avoids both reference lengths and a separately trained length predictor.

Our approach is to ask the frozen backbone which canvas it is most prepared to fill. Given a source sentence, \method (EV) forms a small set of candidate canvas lengths, runs one all-mask forward pass for each candidate, and scores the canvas by mean predictive entropy. EV then decodes on the canvas with the lowest mean entropy. The method adds no trainable parameters and does not use reference lengths, development set tuning, or an external length predictor. Its target is not exact reference length prediction; it chooses a length that the masked diffusion model can denoise well.

We make four contributions in this LLaDA-8B LoRA-SFT masked diffusion MT setting.
\begin{enumerate}[leftmargin=*,topsep=2pt,itemsep=1pt]
  \item We identify target length selection as a measurable bottleneck in fixed canvas masked diffusion MT and include a controlled comparison with decoding-order choices under matched budgets.
  \item We introduce \method, which requires no additional training and uses entropy on all-mask canvases to choose a canvas the backbone can denoise well.
  \item We show across \mbox{En$\to$Zh}, \mbox{Zh$\to$En}, and \mbox{En$\to$De} that EV recovers $64.9\%$, $65.3\%$, and $33.0\%$ of the COMET-22 improvement obtained by supplying reference target lengths, while not simply matching those lengths.
  \item We test the length selection finding with statistical tests, bidirectional expert human evaluation, coverage and error type analysis, a matched data LLaMA-3-8B AR baseline, reveal-order controls, and cross backbone checks on Dream-Base and DiffuLLaMA.
\end{enumerate}

\section{Related Work}
\label{sec:related}

\paragraph{Masked diffusion decoding.} Text diffusion models range from continuous-embedding formulations \citep{li2022diffusionlm,gong2023diffuseq} to discrete absorbing-state and masked objectives \citep{austin2021d3pm,lou2024sedd,sahoo2024mdlm}. Recent 8B masked diffusion LMs such as LLaDA and Dream report strong open-ended generation and reasoning results \citep{nie2025llada,ye2025dream}. In the fixed-canvas MT setting studied here, the target length is supplied before denoising; when this length is unsuitable, coverage and redundancy change before the unmasking order is chosen.

\paragraph{Length selection in machine translation.} Parallel MT systems have long treated length as an explicit modeling choice. Early NAT predicts target length before parallel decoding \citep{gu2017nonautoregressive}; CMLM adds a trained length head and small-beam length search \citep{ghazvininejad2019mask}; GLAT changes the training objective \citep{qian2021glancing}; Levenshtein Transformer avoids a fixed canvas through insertion and deletion operations \citep{gu2019levenshtein}. \citet{wang-etal-2021-length} show that the best candidate from an oracle CMLM length beam can outperform decoding at the exact reference length. These methods are effective when the architecture and objective are built around them. Our setting keeps the pretrained decoder-only masked diffusion backbone fixed after LoRA-SFT, so EV selects length at test time from the model's all-mask entropy rather than adding a length predictor, length beam, or edit-based decoder.

\paragraph{Variable length diffusion language models.} Recent work handles unknown generation length in several ways. DAEDAL adjusts an initial length before denoising and can insert masks during denoising \citep{li2025fixedtrainingfreevariablelengthdenoising}; $\rho$-EOS expands or contracts the sequence during denoising using EOS density \citep{yang2026rhotexttteostrainingfreebidirectionalvariablelength}. CAL searches candidate lengths before decoding using calibrated confidence from the first denoising step \citep{liu2026diffusionlmsapproximateoptimal}, while LR-DLLM corrects length bias in confidence scores across candidate lengths \citep{cheng2026improvingvariablelengthgenerationdiffusion}. SmartCrop estimates the output length from the prompt and crops the canvas before generation \citep{rossi2026diffusionlanguagemodelsnatively}. FlexMDM changes training and generation to support mask insertion \citep{kim2025anyorderflexiblelengthmasked}; dLLM-Var trains the model to produce EOS and outputs of varying length \citep{yang2025diffusionllmnativevariable}. EV uses mean entropy from all mask predictions to choose a canvas for each source sentence before denoising. It does not require a learned length head, selector training, or changes to the denoising schedule.

\paragraph{Autoregressive LLMs for MT.} Recent AR LLM-MT systems improve translation through MT-specific continued training, instruction tuning, or preference optimization \citep{xu2023paradigm,xu2024contrastive,alves2024tower}. We use a matched-data LLaMA-3-8B LoRA-SFT baseline to calibrate the masked-diffusion setting against this line of work, but EV targets a different bottleneck: choosing the fixed canvas before denoising.

\paragraph{Reveal order methods.} A separate line of masked generation work studies which tokens to reveal, not which canvas to reveal them on. MaskGIT uses confidence-based parallel unmasking \citep{chang2022maskgit}; OeMDM/LoMDM learn generation orders with the backbone \citep{hong2026oemdm}; LogicDiff designs task-specific orders for reasoning \citep{aman2026logicdiff}. These methods are not direct MT length-selection baselines, so we use them to locate EV relative to order scheduling. Our controlled diagnostic asks the complementary question for MT: under matched budgets, how much variation comes from the canvas length source versus the tested unmasking orders (\S\ref{sec:analysis-length-order}). EV can be used with standard minimum-entropy decoding, which reveals lower-entropy positions first, because it targets this separate length-selection bottleneck.

\section{Method: \method}
\label{sec:method}

\subsection{Overview}
\label{sec:method-overview}

\begin{figure*}[t]
  \centering
  \includegraphics[width=\textwidth]{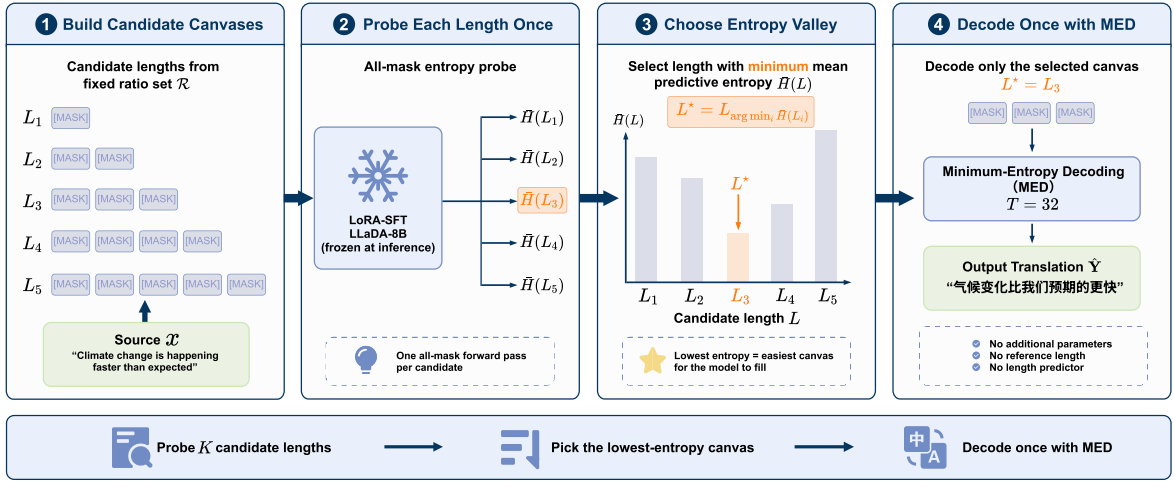}
  \caption{\textbf{Entropy-Valley method overview.}
Given a source sentence, EV uses five candidate ratios from the fixed set $\Rset$, yielding up to five distinct target canvases after duplicate lengths are removed. It then probes each candidate length once with an all-mask forward pass through the frozen LoRA-SFT LLaDA-8B model and records the mean predictive entropy $\Hbar(L)$ over all slots except the designated EOS slot. EV selects the entropy valley, $L^\star=\arg\min_L \Hbar(L)$, and decodes only this selected canvas with minimum-entropy decoding (MED) for $T{=}32$ steps. The procedure adds no parameters, does not use the reference length of the current sentence, and requires no length predictor.}
  \label{fig:method}
\end{figure*}

Given a source sentence $\xvec$, our goal is to choose a target canvas length before masked diffusion decoding begins. The output of the length selector is $\Lstar$; the translation is then produced by the same minimum-entropy decoding (MED) schedule used by the baseline, only on the selected canvas. As shown in Figure~\ref{fig:method}, \method (EV) follows four operations. First, EV builds a small set of candidate canvas lengths from a fixed source-to-target ratio set. Second, it probes each candidate length once with an all-mask forward pass and scores the canvas by mean predictive entropy. Third, EV selects the lowest-entropy canvas, or entropy valley, and then runs standard MED decoding on that canvas. EV changes only the target length supplied to the decoder, not the backbone, training loss, or unmasking schedule.

\subsection{Problem formulation}
\label{sec:method-setup}

A masked diffusion language model receives a source prompt and a target canvas $\yvec \in (\V \cup \{\Mask\})^L$. It denoises the canvas over $\Tsteps$ steps under an order schedule $\pi$. In our main experiments, $\pi$ is MED, which reveals lower-entropy positions first. Training appends EOS to each target sequence. At inference, the allocated canvas includes a final slot designated for EOS, and the decoded string is truncated at the first EOS. The canvas length $L$ must be supplied before decoding.

We study test-time length selection for a frozen LoRA-SFT masked diffusion MT system. A length selector takes $\xvec$ and returns $\Lhat(\xvec)\in\mathbb{N}_+$ from a finite candidate set
\[
\Cset(\xvec)=\{\max\{1,\lfloor r|\xvec|\rfloor\}+1 : r\in\Rset\}.
\]
Here $|\xvec|$ is the source-side tokenizer length under the same tokenizer used by the backbone, and the final $+1$ reserves the designated EOS slot. At inference, the selector does not use the reference length of the current sentence or a separately trained length predictor. It scores the predefined candidates using the frozen model's all-mask uncertainty:
\[
\begin{aligned}
\Lstar &= \argmin_{L\in\Cset(\xvec)} \Hbar(L),\\
\Hbar(L) &= \frac{1}{L-1}\sum_{i=1}^{L-1}
\Hent\!\left(p_\theta(y_i\mid \xvec,\Mask^L)\right).
\end{aligned}
\]
The target is not to recover $\Lref$. EV chooses the candidate length that the current backbone appears most prepared to denoise.

\subsection{Candidate canvas construction}
\label{sec:method-R}

A fixed ratio is cheap, but it assigns the same compression rule to short commands, entity-heavy sentences, and idiomatic translations. Probing every possible length would require no additional training, but it would spend computation on canvases that are implausible for the language pair. EV instead searches a small set of plausible sentence-level lengths.

The median ratios below are WMT19 training corpus statistics and serve as reference scales. The ratio grid is set separately for each direction. Diagnostic range comparisons on WMT22 subsets informed the reported compact grids. Once chosen, each grid was used for every sentence and decoding method in that direction:

\begin{center}
\small
\setlength{\tabcolsep}{1.9pt}
\begin{tabular}{l c l}
\toprule
\textbf{Direction} & \textbf{Median ratio} & \textbf{$\Rset$} \\
\midrule
En$\to$Zh & 0.80 & $\{0.70, 0.75, 0.80, 0.85, 0.90\}$ \\
Zh$\to$En & 1.24 & $\{1.00, 1.10, 1.20, 1.30, 1.40\}$ \\
En$\to$De & 1.48 & $\{1.50, 1.60, 1.70, 1.80, 1.90\}$ \\
\bottomrule
\end{tabular}
\end{center}

\noindent For a source sentence $\xvec$, EV maps the ratios to canvas lengths using the formula above and removes duplicates. The five ratios can thus yield fewer than five distinct lengths for short source sentences. These grids provide a compact set of plausible canvases, while selection for each sentence comes from the frozen model's all-mask entropy.

The candidate set gives EV only one role at test time: choose among a few plausible canvases for the current source. It does not use the reference length of the current sentence or any manual adjustment during decoding. The range still matters; Appendix Table~\ref{tab:candidate_range} reports the candidate-width control.

\subsection{Entropy-Valley scoring}
\label{sec:method-intuition}

The decoder must choose $L$ before generation. A useful score should be available then and should come from the same model that fills the canvas. On an all-mask canvas, a too-short length forces source content into too few slots, while a too-long length leaves positions with diffuse alternatives. Mean predictive entropy exposes this mismatch without running the full denoising loop.

For each $L\in\Cset(\xvec)$, EV runs one forward pass on $[\,\text{prompt}(\xvec)\,]\Vert\Mask^L$ and computes $\Hbar(L)$. The final slot is excluded because it is designated for EOS and typically has entropy close to zero. Averaging over the remaining $L-1$ slots compares candidates of different lengths without allowing this slot to dominate the score. EV selects the entropy valley, $\Lstar=\argmin_{L\in\Cset(\xvec)}\Hbar(L)$.

The score makes EV a denoising-friendly canvas selector rather than a reference-length regressor. The chosen length need not match the reference exactly if it gives the backbone a canvas it can fill reliably (details in \S\ref{sec:analysis-dissoc}).

\subsection{Decoding algorithm}
\label{sec:method-algo}

Algorithm~\ref{alg:ev} gives the full inference procedure. The only difference from the fixed-ratio baseline is the loop over candidate lengths. Once $\Lstar$ is selected, decoding uses the standard MED schedule for the same $\Tsteps$ steps used by the baseline.

\begin{algorithm}[H]
\small
\caption{\textbf{\method length selection and decoding}}
\label{alg:ev}
\begin{algorithmic}[1]
\Require Source sentence $\xvec$; frozen model $p_\theta$; fixed ratio set $\Rset$; decoding steps $\Tsteps$; MED schedule $\pi$.
\Ensure Translation decoded from the selected canvas.
\State Form candidate lengths $\Cset(\xvec) \gets \{\max\{1,\lfloor r|\xvec|\rfloor\}+1 : r \in \Rset\}$; deduplicate.
\For{$L \in \Cset(\xvec)$}
  \State Build the all-mask input $[\,\text{prompt}(\xvec)\,] \,\Vert\, \Mask^L$.
  \State Record $q_1^L,\ldots,q_L^L$ with one forward pass.
  \State Set $\Hbar(L)$ to the mean entropy of $q_1^L,\ldots,q_{L-1}^L$.
\EndFor
\State Select $\Lstar \gets \argmin_{L\in\Cset(\xvec)} \Hbar(L)$.
\State Decode $[\,\text{prompt}(\xvec)\,] \,\Vert\, \Mask^{\Lstar}$ with MED schedule $\pi$ for $\Tsteps$ steps.
\State Return the decoded target string, truncated at the first EOS as in the shared decoding protocol.
\end{algorithmic}
\end{algorithm}

\subsection{Cost and implementation details}
\label{sec:method-cost}

\method uses five ratios and probes at most five distinct candidate lengths before decoding. At the default $T{=}32$, this adds at most five probe passes to the 32 denoising passes used by the baseline; fewer are needed when multiple ratios map to the same integer length. Candidates can be evaluated sequentially with no extra peak memory, as in our setup, or batched if memory permits. EV adds no trainable parameters, alignment model, or separately trained length predictor. Appendix Table~\ref{tab:fwdcount_control} controls for the extra forward pass budget, and Appendix~\ref{sec:appendix_method} records the canvas conventions, external length property, and length versus order protocol.

\section{Experiments}
\label{sec:exp}

The experiments test six claims about \method: the length problem is measurable, EV improves over a fixed corpus ratio, the gain is not just reference-length matching, the candidate design matters, automatic gains align with human judgments on \mbox{En$\leftrightarrow$Zh}, and the method has clear boundary cases across language pairs and backbones. We first fix the protocol, then report the main LLaDA results, matched-data AR calibration, cross-backbone scope, analysis and controls, human evaluation, and a brief discussion of what EV selects. Additional paired significance tests are in Appendix~\ref{sec:appendix_ablations}.

\subsection{Setup}
\label{sec:exp-setup}

We LoRA-SFT LLaDA-8B-Base \citep{nie2025llada} on $200$k WMT19 training pairs per direction (\mbox{En$\leftrightarrow$Zh} and \mbox{En$\to$De}), train three independent runs for each direction, and evaluate on all $2{,}037$ WMT22 samples per direction with COMET-22 \citep{rei2022comet} and sacreBLEU \citep{post2018call}. Decoding uses $\Tsteps{=}32$ MED with EOS truncation; full protocol, significance procedure, hyperparameters, and compute are in Appendix~\ref{sec:appendix_repro}.

\subsection{Main Results}
\label{sec:exp-main}

\begin{figure*}[t]
  \centering
  \includegraphics[width=\textwidth]{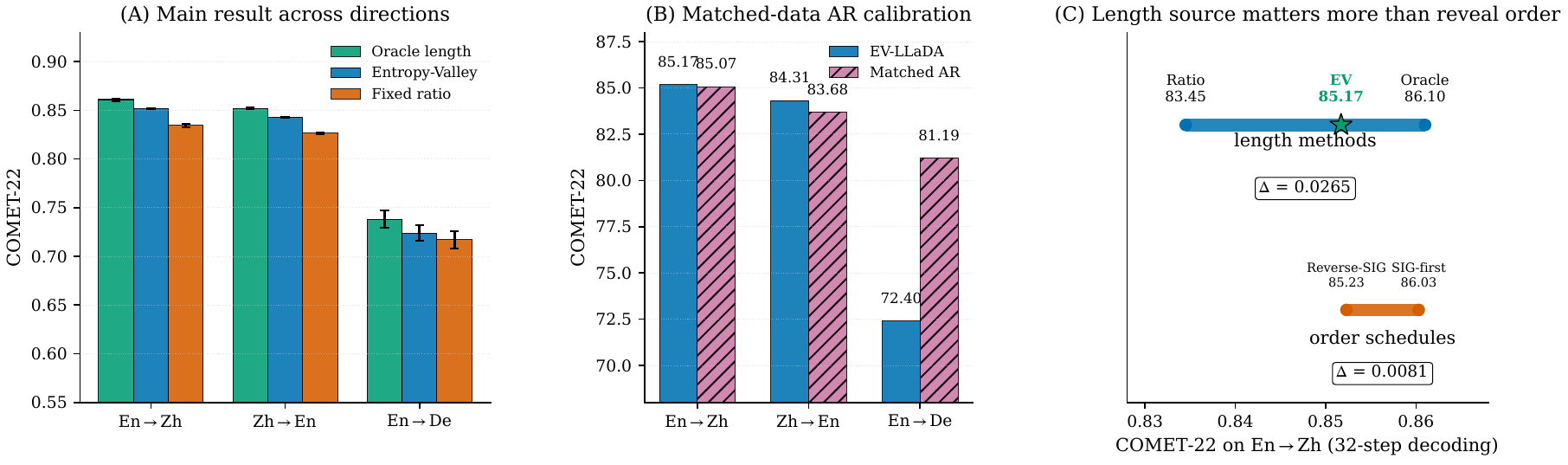}
  \caption{\textbf{Length-adaptive decoding for masked-diffusion MT.}
(A) Mean COMET-22 over three runs for a reference-length upper bound, EV, and a corpus-ratio baseline across three directions. EV closes $64.9\%$ on \mbox{En$\to$Zh}, $65.3\%$ on \mbox{Zh$\to$En}, and $33.0\%$ on \mbox{En$\to$De}.
(B) Matched-data AR calibration across three directions.
(C) Controlled En$\to$Zh diagnostic: with reference target lengths fixed, the tested source-guided unmasking schedules span $0.0081$ COMET. With the schedule fixed to MED, three length choices span $0.0265$ COMET. The order-schedule sweep uses reference lengths, so it is a diagnostic rather than a deployable alternative to EV.}
  \label{fig:hero}
\end{figure*}

\begin{table*}[t]
\centering
\footnotesize
\setlength{\tabcolsep}{4pt}
\begin{tabular}{ll cc cc}
\toprule
 & & \multicolumn{2}{c}{\textbf{COMET-22}} & \multicolumn{2}{c}{\textbf{sacreBLEU}} \\
 \cmidrule(lr){3-4}\cmidrule(lr){5-6}
\textbf{Direction} & \textbf{Method} & Mean $\pm$ Std & Gap Closure & Mean $\pm$ Std & Gap Closure \\
\midrule
\multirow{3}{*}{En$\to$Zh}
 & Oracle length  & $0.8610 \pm 0.0013$ & 100\% & $40.81 \pm 0.23$ & 100\% \\
 & Fixed ratio 0.8 & $0.8345 \pm 0.0017$ & 0\% (baseline) & $36.72 \pm 0.22$ & 0\% \\
\rowcolor{EVRow}
 & \textbf{Entropy-Valley} & $\mathbf{0.8517 \pm 0.0006}$ & $\mathbf{64.9 \pm 7.4\%}$ & $\mathbf{38.57 \pm 0.13}$ & 45.3\% \\
\midrule
\multirow{3}{*}{Zh$\to$En}
 & Oracle length  & $0.8519 \pm 0.0007$ & 100\% & $27.93 \pm 0.23$ & 100\% \\
 & Fixed ratio 1.2 & $0.8266 \pm 0.0010$ & 0\% (baseline) & $23.65 \pm 0.21$ & 0\% \\
\rowcolor{EVRow}
 & \textbf{Entropy-Valley} & $\mathbf{0.8431 \pm 0.0004}$ & $\mathbf{65.3 \pm 0.8\%}$ & $\mathbf{25.28 \pm 0.33}$ & 38.1\% \\
\midrule
\multirow{3}{*}{En$\to$De}
 & Oracle length  & $0.7382 \pm 0.0090$ & 100\% & $22.55 \pm 0.77$ & 100\% \\
 & Fixed ratio 1.8 & $0.7170 \pm 0.0090$ & 0\% (baseline) & $20.73 \pm 0.68$ & 0\% \\
\rowcolor{EVRow}
 & \textbf{Entropy-Valley} & $\mathbf{0.7240 \pm 0.0078}$ & $\mathbf{33.0 \pm 8.4\%}$ & $\mathbf{21.55 \pm 0.85}$ & 45.1\% \\
\bottomrule
\end{tabular}
\caption{\textbf{Main LLaDA-8B+LoRA results on WMT22 with 32-step MED decoding.} Values are mean $\pm$ std over three runs. \oracleref uses the reference target length at decode time and is an upper bound. Gap closure is $(\mathrm{method}-\mathrm{Ratio})/(\mathrm{Oracle}-\mathrm{Ratio})$. Paired tests are in Appendix Table~\ref{tab:full_significance}.}
\label{tab:main}
\end{table*}

Figure~\ref{fig:hero}A gives the COMET overview, and Table~\ref{tab:main} reports the corresponding COMET and sacreBLEU values. With the backbone, training data, and MED schedule fixed, the comparison isolates the target-canvas selector. Replacing the training-corpus \ratio with \method raises COMET-22 by $+0.0172$ on \mbox{En$\to$Zh} and $+0.0165$ on \mbox{Zh$\to$En}, closing $64.9{\pm}7.4\%$ and $65.3{\pm}0.8\%$ of the COMET gap between the length oracle and the fixed-ratio baseline. On \mbox{En$\to$De}, EV gains $+0.0070$ COMET and closes $33.0{\pm}8.4\%$ of the same gap. The corresponding sacreBLEU gains are $+1.85$, $+1.63$, and $+0.82$, respectively.

The paired tests give the clearest support on \mbox{En$\leftrightarrow$Zh}, where EV remains above \ratio under both paired bootstrap and Wilcoxon tests (Appendix Table~\ref{tab:full_significance}). For \mbox{En$\to$De}, the mean over three runs is positive, but the sentence-level evidence is weaker. This direction is a boundary case for length selection within the fixed LLaDA system.

The extra EV probes do not explain the main gains.
Appendix \mbox{Table~\ref{tab:fwdcount_control}} gives \ratio the same or more forward passes and recovers at most $0.001$ COMET, far below the \mbox{En$\leftrightarrow$Zh} gains in \mbox{Table~\ref{tab:main}}.

\subsection{Reference and matched-data baselines}
\label{sec:exp-ar}

\begin{table*}[t]
\centering
\footnotesize
\setlength{\tabcolsep}{5pt}
\begin{tabular}{l r l r r r}
\toprule
\textbf{System} & \textbf{Size} & \textbf{Setting} & \textbf{En$\to$Zh} & \textbf{Zh$\to$En} & \textbf{En$\to$De} \\
\midrule
\rowcolor{EVGroup}
\multicolumn{6}{l}{\textit{LLaDA-8B masked diffusion}} \\
Fixed ratio                  & 8B  & LoRA-SFT             & 83.45 & 82.66 & 71.70 \\
\rowcolor{EVRow}
\textbf{Entropy-Valley}      & 8B  & LoRA-SFT             & \textbf{85.17} & \textbf{84.31} & 72.40 \\
Oracle length$^{\dagger}$    & 8B  & LoRA-SFT             & 86.10 & 85.19 & 73.82 \\
\midrule
\rowcolor{EVGroup}
\multicolumn{6}{l}{\textit{Matched-data autoregressive baseline}} \\
\textbf{LLaMA-3-8B + LoRA-SFT} & 8B & LoRA-SFT & $85.07{\pm}0.03$ & $83.68{\pm}0.07$ & $\mathbf{81.19{\pm}0.11}$ \\
\bottomrule
\end{tabular}
\caption{\textbf{Matched-data COMET-22 calibration on WMT22.} LLaDA rows show the Table~\ref{tab:main} means over three runs scaled $\times 100$; the matched AR row reports mean$\pm$std over three runs after training on the same fine-tuning data and evaluating on the same test sets. $^{\dagger}$Oracle length supplies the reference target length at decode time and is an upper bound, not a deployable method.}
\label{tab:ar_reference}
\end{table*}

Figure~\ref{fig:hero}B and Table~\ref{tab:ar_reference} use a matched autoregressive system as a calibration point, not as a target-canvas baseline. With the same fine-tuning data, LLaDA with \method is essentially tied with LLaMA-3-8B + LoRA-SFT on \mbox{En$\to$Zh} and is higher on \mbox{Zh$\to$En}. This calibrates the masked-diffusion setting without changing the claim: EV addresses canvas selection within a fixed backbone.

\mbox{En$\to$De} gives the boundary. The matched autoregressive system is about $7.4$ COMET points above the LLaDA \oracleref row and about $8.8$ points above LLaDA with \method. The gap remains even when the reference target length is supplied, so the \mbox{En$\to$De} shortfall is not mainly a target-canvas selection error. EV is scoped to fixed-backbone canvas selection.

\subsection{Scope across masked-diffusion backbones}
\label{sec:exp-crossbb}

\begin{figure}[t]
\centering
\includegraphics[width=\columnwidth]{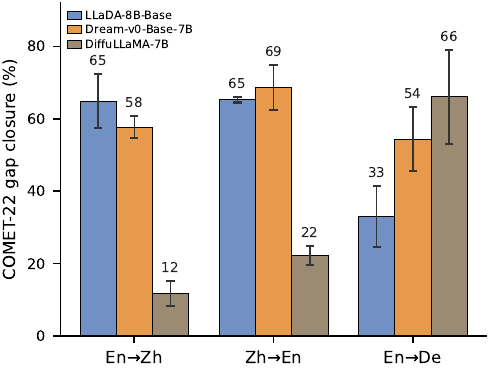}
\caption{\textbf{Cross-backbone scope check using COMET-22 gap closure (\%) for three masked-diffusion backbones and three directions.} Gap closure is measured between \ratio and \oracleref for each backbone--direction pair. Bars show means over three runs; error bars show standard deviations across runs. Full COMET-22 and sacreBLEU values are in Appendix~\ref{sec:appendix_crossbb} (Tables~\ref{tab:crossbb} and~\ref{tab:crossbb_bleu}).}
\label{fig:crossbb}
\end{figure}

We repeat the \oracleref, \ratio, and \method protocol on Dream-Base and DiffuLLaMA to test if the length-selection pattern is specific to LLaDA. These runs use the same $200$k SFT data scale, MED decoding protocol, and three directions as the main evaluation. Figure~\ref{fig:crossbb} summarizes the cross-backbone gap-closure pattern.

Across all tested directions and backbones, \method remains above \ratio, but the amount of oracle-gap closure changes by model family. This supports the entropy-based length criterion beyond LLaDA while keeping the claim scoped to the tested systems.

The variation is not a causal attribution. Dream-Base and DiffuLLaMA change tokenizer, initialization, pretraining mixture, and architecture at the same time. Their larger En$\to$De closures do not identify which factor helps, and DiffuLLaMA's smaller Chinese-side closures are consistent with limited Chinese tokenizer coverage.

\subsection{Controls for the EV gain}
\label{sec:exp-ablations}

\noindent \method changes only the target-canvas choice. The controls below ask whether the main gains instead come from reveal order, reference-length matching, candidate-window tuning, a better global ratio, or metric-only artifacts.

\paragraph{Length choice versus reveal order.}\label{sec:analysis-length-order}
The order diagnostic fixes the target length to the reference length and changes only the reveal schedule. On \mbox{En$\to$Zh}, the EV--Ratio difference is larger than the full span across the tested reveal orders. Strict left-to-right and random order also score below MED. In this controlled setting, target canvas choice produces the larger variation among the evaluated decoding decisions, while poor reveal orders can still hurt (Appendix Table~\ref{tab:order_sweep_summary}).

\begin{figure}[t]
  \centering
  \includegraphics[width=\columnwidth]{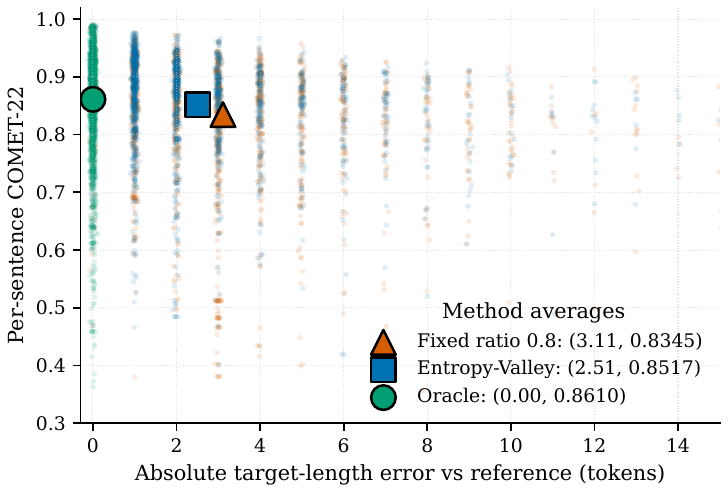}
  \caption{\textbf{Quality need not track reference length.}
Target length error and COMET-22 on En$\to$Zh WMT22 ($N{=}2037$). Points show sentence scores; MAE values use the same outputs, and centroid heights use the main results in Table~\ref{tab:main}.}
  \label{fig:dissociation}
\end{figure}

\paragraph{Not reference matching.}\label{sec:analysis-dissoc}
EV does not aim to predict the reference length. Figure~\ref{fig:dissociation} compares target length error with COMET-22 on \mbox{En$\to$Zh}. EV remains far from \oracleref in length error, yet recovers about $65\%$ of the \oracleref--\ratio COMET gap. This pattern supports the method framing: entropy selects a canvas the backbone can denoise well, rather than a human reference length.

\paragraph{Candidate range.} The candidate range defines what EV can choose. Appendix Table~\ref{tab:candidate_range} shows that wider windows are not uniformly better, supporting a compact fixed window rather than unrestricted length search.

\paragraph{Fixed ratio sweep.} EV is not just a better global ratio. Appendix Table~\ref{tab:ratio_sweep} sweeps fixed ratios around each corpus median on 500-sentence diagnostic subsets. EV remains above the best fixed ratio in all three directions. A single ratio applies the same compression rule to every source sentence; EV changes the canvas per sentence, which is why retuning one ratio for the whole corpus does not reproduce the gain.

\paragraph{Multi-candidate decoding.} A heavier decode-many-then-score alternative does not explain the result. Appendix Table~\ref{tab:multi_candidate} decodes several neighboring canvas lengths to completion and then selects one by average log-probability. Unlike EV's up to five one-pass probes (\S\ref{sec:method-cost}), this alternative pays for multiple completed decodes and is lower-scoring in the tested setting.

\paragraph{Direct diffusion length baselines.}
On independent runs for each direction using the same translation setup, EV exceeds DAEDAL \citep{li2025fixedtrainingfreevariablelengthdenoising} in both directions and CAL \citep{liu2026diffusionlmsapproximateoptimal} on \mbox{En$\to$Zh}. EV and CAL are comparable on \mbox{Zh$\to$En}. EV uses lower measured inference cost than CAL in this setup (Appendix Table~\ref{tab:direct_length_baselines}).

\paragraph{Coverage and limits.}\label{sec:analysis-faithfulness}
The automatic gains align with source coverage, not only with average COMET. Table~\ref{tab:faithfulness_summary} focuses on diagnostics tied to the length-selection story. On \mbox{En$\to$Zh}, EV preserves more placeholders and numbers than \ratio. These items are sensitive to canvas length because a short canvas can drop literal content before reveal order matters.

\begin{table}[t]
\centering
\footnotesize
\setlength{\tabcolsep}{4pt}
\begin{tabular}{l r r r}
\toprule
\textbf{Diagnostic} & \textbf{N} & \textbf{Ratio} & \textbf{EV} \\
\midrule
Placeholder retention & 58 & 66.9\% & \textbf{89.1\%} \\
Number retention & 290 & 77.6\% & \textbf{81.3\%} \\
$r{\ge}0.8$ COMET & 1005 & 0.8056 & \textbf{0.8402} \\
$r{<}0.6$ COMET & 236 & 0.8449 & \textbf{0.8503} \\
\bottomrule
\end{tabular}
\caption{\textbf{Coverage and length buckets on En$\to$Zh.} Retention rows measure whether literal source items such as placeholders and numbers are preserved. COMET rows group sentences by $r$, the reference target length divided by the source length under the LLaDA tokenizer. Appendix~\ref{sec:appendix_coverage} gives the definitions, confidence intervals, and Zh$\to$En companion rows.}
\label{tab:faithfulness_summary}
\end{table}

The same buckets show the current boundary. The largest COMET gain appears in the dominant $r{\ge}0.8$ bucket, where the default candidate grid is well aligned. The $r{<}0.6$ bucket still improves on average, but individual failures can occur when the needed compression lies below the grid. Figure~\ref{fig:cases} shows such a case, and Appendix~\ref{sec:appendix_cases} gives the full failure breakdown.

\subsection{Human evaluation on \texorpdfstring{\mbox{En$\leftrightarrow$Zh}}{En<->Zh}}
\label{sec:exp-human}

\begin{table}[t]
\centering
\footnotesize
\setlength{\tabcolsep}{2.45pt}
\begin{tabular}{l c c c}
\toprule
\textbf{Direction} & $\Delta$Adequacy & $\Delta$Fluency & \textbf{Preference} \\
\textbf{} & \textbf{EV $-$ Ratio} & \textbf{EV $-$ Ratio} & \textbf{EV / Ratio / Tie} \\
\midrule
\mbox{En$\to$Zh} & $+0.18$ & $+0.18$ & 28 / 19 / 53 \\
\mbox{Zh$\to$En} & $\mathbf{+0.50}$ & $+0.11$ & \textbf{45 / 25 / 30} \\
\bottomrule
\end{tabular}
\caption{\textbf{Human evaluation on $100$ stratified WMT22 sentences per direction.} Score differences are EV $-$ Ratio, averaged over three bilingual expert annotators; preference is the per-sentence majority vote.}
\label{tab:human_eval}
\end{table}

The human study checks whether the automatic \mbox{En$\leftrightarrow$Zh} gains are also reflected in human adequacy judgments.
Three experts bilingual in Chinese and English compared anonymised \method and \ratio outputs.
We use $100$ stratified WMT22 sentences for En$\rightarrow$Zh and another $100$ for Zh$\rightarrow$En, with the source and gold translation shown.
The full protocol, rater statistics, and metric correlation results are reported in \mbox{Appendix~\ref{sec:appendix_human_protocol}}.

Human judgments follow the automatic pattern, with clearer evidence on \mbox{Zh$\to$En}. The main signal is adequacy rather than fluency: EV raises adequacy by $+0.50$ on \mbox{Zh$\to$En} and wins more majority preferences than \ratio ($45$ vs.\ $25$). On \mbox{En$\to$Zh}, the adequacy margin is smaller ($+0.18$), and most sentences are ties under majority preference. Fluency changes are small in both directions. This pattern matches the length-selection story: EV helps preserve source content rather than improving fluency across the board.

\subsection{Discussion: what Entropy-Valley selects}
\label{sec:discussion}

The diagnostics identify EV as a canvas selector, not a reference-length predictor. Figure~\ref{fig:dissociation} shows that high translation quality does not require matching the reference length exactly. Table~\ref{tab:faithfulness_summary} gives the complementary mechanism: when a fixed ratio allocates too few slots, the selected canvas can preserve source-side placeholders and numbers that would otherwise be dropped. The useful signal is the backbone's uncertainty about how well it can fill a candidate canvas before decoding begins.

The boundary comes from the candidate set.
The fixed ratio set keeps EV free from extra training, but it also limits the canvases EV can choose at inference time.
If the needed compression falls outside this set, as in Figure~\ref{fig:cases}, the entropy score cannot select the missing canvas.
Figure~\ref{fig:crossbb} shows a second boundary: EV can select a better canvas for each tested backbone, but gains remain bounded by the backbone and tokenizer.

\section{Conclusion}
\label{sec:conclusion}

This paper identifies target-canvas selection as a structural bottleneck in fixed-canvas masked-diffusion MT. The issue is not only how to reveal masked positions, but which canvas the decoder is asked to fill before denoising begins. A short canvas can drop source content; an overlong canvas can create unsupported slots. The central takeaway is that masked-diffusion MT needs an explicit test-time length decision.

\method turns this decision into a model-internal query. By probing a small set of all-mask canvases, EV chooses the candidate the backbone appears most prepared to denoise, without adding a length head or using reference lengths. Across the tested directions, this simple selector improves over a fixed corpus ratio, aligns with human adequacy judgments on \mbox{En$\leftrightarrow$Zh}, and preserves literal source content that short canvases can lose. In the controlled \mbox{En$\to$Zh} comparison, target canvas choice produces the larger variation among the evaluated decoding decisions, while strict left-to-right and random reveal orders also reduce quality.

\section*{Limitations}
\label{sec:limitations}

Our deepest evaluation uses LLaDA-8B-Base. Dream-Base and DiffuLLaMA extend the same protocol to two additional backbones, and EV remains above \ratio in every tested cell. Absolute scores and oracle-gap closures still vary with the backbone and tokenizer, so broader generalization requires more model families and language pairs.

\mbox{En$\to$De} is the clearest boundary for the current LLaDA system. EV-LLaDA improves over \ratio, but the sentence-level evidence is weaker than on \mbox{En$\leftrightarrow$Zh} (Appendix~\ref{sec:appendix_ablations}). The matched AR baseline is about $8.8$ COMET points above EV-LLaDA and about $7.4$ points above even the LLaDA length oracle on the $0$--$100$ scale, so the \mbox{En$\to$De} gap is not mainly a target-canvas selection error. The step-budget sweep gives the same warning from another angle: at $\Tsteps{\ge}64$, the fixed \mbox{En$\to$De} ratio slightly exceeds EV (Appendix Figure~\ref{fig:pareto} and Table~\ref{tab:full_pareto}).

The human evaluation is limited to \mbox{En$\leftrightarrow$Zh}. These judgments support the automatic adequacy gains most clearly on \mbox{Zh$\to$En}, while \mbox{En$\to$Zh} is positive but weaker under preference testing. We did not run human evaluation for \mbox{En$\to$De} or De$\to$Fr, so conclusions for those directions rely on automatic metrics; the \mbox{En$\to$De} analysis also includes matched AR calibration.

The fixed candidate ratio set is another intended constraint. The reported main grids are direction specific and were informed by comparisons on WMT22 diagnostic subsets. The De$\to$Fr transfer uses one fixed grid without changes, but broader transfer of the procedure for choosing the grid remains open. The fixed set keeps EV training-free and prevents an unrestricted length search, but it also limits outlier sentences whose target length requires stronger compression. With $\Rset{=}\{0.70,\ldots,0.90\}$ on \mbox{En$\to$Zh}, EV cannot choose a canvas below $0.70|\xvec|$; Appendix~\ref{sec:appendix_cases} gives such a failure case. A natural next step is a compression-aware candidate generator that expands or shifts $\Rset$ when the entropy curve suggests that the fixed window is insufficient. \label{sec:limitations-candidate-grid}

\section*{Ethical Considerations}
\label{sec:ethics}

EV changes only inference-time canvas selection and does not alter the backbone or training objective. The resulting system still inherits risks from the backbone, translation fine-tuning data, and decoding procedure. Because backbone pretraining data and tokenization vary across model families and scales, deployment should evaluate fairness and robustness for the intended language pairs and user domains. Automatic quality gains do not remove errors in individual sentences, so use in medical, legal, and other high risk settings should retain human review.

We use public WMT news benchmarks and evaluation packages under their public research terms or licenses. WMT news data may contain names or sensitive events already present in public news text. We release the processed experiment datasets and trained LoRA adapters described in Appendix~\ref{sec:appendix_repro}. The datasets remain subject to the original WMT research terms, and the adapters retain the applicable base model license. Backbone weights are obtained from their original providers. We do not collect private user data or attempt to identify individuals.

The three evaluators were professional translators bilingual in Chinese and English. They rated both \mbox{En$\to$Zh} and \mbox{Zh$\to$En} batches using public WMT22 sentences and outputs with system identities hidden. They were informed of the study purpose, worked independently, consented to the use of ratings without personal identifiers and aggregate statistics for research reporting, and were compensated at local professional translator hourly rates. The task collected no sensitive personal data and, under the rules of our institution, did not require formal ethics review.

\FloatBarrier

\bibliography{references}

\clearpage

\appendix

\section{Reproducibility and Configuration}
\label{sec:appendix_repro}

Unless stated otherwise, the LLaDA experiments use the following configuration. Subset analyses that change $N$ are identified at the relevant table.

\paragraph{Backbone.} LLaDA-8B-Base \citep{nie2025llada}; Transformer, $8.02$B parameters. We use the released weights without modification. LoRA \citep{hu2022lora} targets seven modules per Transformer block: the attention projection matrices $\{q, k, v, o\}$ and the FFN matrices $\{\textrm{ff\_proj}, \textrm{up\_proj}, \textrm{ff\_out}\}$ (no adaptation of embeddings): rank $64$, $\alpha{=}128$, dropout $0.05$. Trainable parameters $\approx 157$M ($1.95\%$ of the backbone). bf16 training and inference throughout.

\paragraph{Training.} AdamW optimiser ($\beta_1{=}0.9$, $\beta_2{=}0.95$, weight decay $0.01$); cosine learning-rate schedule with $5\%$ warm-up, peak LR $2{\times}10^{-4}$; three epochs per direction with a global batch size of $128$ sequences (gradient accumulation ${=}4$ across $8\times$H20 GPUs, per-device micro-batch ${=}4$). Maximum canvas length $1024$ tokens (source prompt $+$ target canvas). Training loss is the standard masked-diffusion cross-entropy with a uniform masking schedule sampled per minibatch.

\begin{table}[H]
\centering
\footnotesize
\setlength{\tabcolsep}{6pt}
\resizebox{\columnwidth}{!}{%
\begin{tabular}{l l}
\toprule
\textbf{Hyperparameter} & \textbf{Value} \\
\midrule
Optimizer & AdamW \\
AdamW betas & $(0.9, 0.95)$ \\
Weight decay & $0.01$ \\
Peak learning rate & $2{\times}10^{-4}$ \\
Scheduler & Cosine decay \\
Warm-up & $5\%$ of training steps \\
Epochs & $3$ per direction \\
Global batch size & $128$ sequences \\
Gradient accumulation & $4$ steps \\
Per-device micro-batch & $4$ sequences \\
Hardware & $8\times$H20-96GB GPUs \\
Precision & bf16 training and inference \\
Maximum canvas length & $1024$ tokens \\
LoRA rank / alpha / dropout & $64$ / $128$ / $0.05$ \\
Model selection & Fixed final checkpoint per run; no EV-specific dev tuning \\
\bottomrule
\end{tabular}
}
\caption{\textbf{Training and adaptation hyperparameters for the LLaDA and matched AR LoRA-SFT runs.}}
\label{tab:hyperparams}
\end{table}

\paragraph{Matched AR baseline.} The matched autoregressive baseline uses LLaMA-3-8B-Base \citep{dubey2024llama3} with LoRA rank $64$, $\alpha{=}128$, dropout $0.05$, and the corresponding seven target module families $\{q,k,v,o,\mathrm{gate},\mathrm{up},\mathrm{down}\}$\_proj. Here matched means that the AR and LLaDA systems use the same SFT pairs, prompt family, LoRA scale, and evaluation data; the module names follow each backbone's architecture. The AR baseline decodes with beam size $4$ and native EOS termination.

\paragraph{Data and runs.} \mbox{En$\leftrightarrow$Zh} uses the HuggingFace \texttt{wmt19/zh-en} training split, and \mbox{En$\to$De} uses \texttt{wmt19/de-en}. We apply only monolingual length filters (English and German: 5--200 whitespace-tokenised words; Chinese: 2--600 characters), then deterministically sample $200$k training pairs and $2$k development pairs per setting. The WMT19 development pairs are not used to set the reported ratio grids or decoding hyperparameters. Three independent training runs per direction yield nine checkpoints. \mbox{En$\to$De} is included as a typologically more distant check, not as a SoTA claim.

\paragraph{Data license and privacy.}
All training and test examples come from public WMT benchmark datasets distributed for machine-translation research. The WMT news benchmarks are public datasets and may contain names or sensitive events already present in public news text. We do not collect private user data, do not attempt to identify individuals, and do not redistribute raw corpora. Human-evaluation forms contain only public WMT sentences and anonymised system outputs.
We use released model weights, evaluation packages, and WMT benchmark data under their respective public research terms or licenses; we do not redistribute the raw WMT corpora or third-party model weights in the public release.

\paragraph{Prompt templates.} Source is formatted as a zero-shot instruction:
\begin{center}\scriptsize
\begin{tabular}{l p{0.62\columnwidth}}
\toprule
Direction & Prompt template \\
\midrule
En$\to$Zh & \texttt{Translate English to Chinese.\textbackslash n\textbackslash nEnglish: \{src\}\textbackslash nChinese: } \\
Zh$\to$En & \texttt{Translate Chinese to English.\textbackslash n\textbackslash nChinese: \{src\}\textbackslash nEnglish: } \\
En$\to$De & \texttt{Translate English to German.\textbackslash n\textbackslash nEnglish: \{src\}\textbackslash nGerman: } \\
\bottomrule
\end{tabular}
\end{center}

\paragraph{Default decoding.} Default decoder is MED at $\Tsteps{=}32$. EOS is not constrained; the decoded string is truncated at the first EOS.

\paragraph{Compared length methods.} The main text compares \oracleref (reference length, upper bound), \ratio (a fixed training-corpus ratio baseline set before evaluation), and \method. The fixed ratios are $0.8$ for \mbox{En$\to$Zh}, $1.2$ for \mbox{Zh$\to$En}, and $1.8$ for \mbox{En$\to$De}. The Pareto sweep uses $\Tsteps \in \{2,4,8,16,32,64,128\}$.

\paragraph{Evaluation.} We evaluate on the WMT22 News Translation test set, $N{=}2037$ per direction. \mbox{En$\to$Zh} and \mbox{Zh$\to$En} use the same WMT22 sentence pairs with source and target roles swapped; \mbox{En$\to$De} uses the WMT22 En--De test set. COMET-22 uses \texttt{unbabel-comet} v2.2 with \texttt{Unbabel/wmt22-comet-da}. BLEU denotes sacreBLEU throughout the paper. The En$\to$Zh signature is {\small\texttt{nrefs:1\allowbreak|case:mixed\allowbreak|eff:no\allowbreak|tok:zh\allowbreak|smooth:exp\allowbreak|version:2.4.0}}; other directions use \texttt{tok:13a}.

\paragraph{Significance.} Paired bootstrap $10$k resamples, Wilcoxon signed-rank (pratt), paired Cohen's $d$.

\paragraph{Compute.} About $420$ H20-96GB GPU-hours for the full experimental grid. At $T{=}32$ on a single H20 with batch size $16$, wall-clock decoding is roughly $1.5$\,s per sentence for \ratio and $1.7$\,s per sentence for \method on \mbox{En$\to$Zh}. The EV probe cap adds at most $15.6\%$ to the forward pass count at $T{=}32$, $7.8\%$ at $T{=}64$, and $3.9\%$ at $T{=}128$; duplicate integer lengths can reduce the realized count.

\paragraph{Resource availability.} We release the implementation, experiment configurations, decoding and evaluation scripts, and human evaluation templates at \url{https://github.com/Entropy-Valley/Entropy-Valley}. The processed experiment datasets and trained LoRA adapters for \mbox{En$\to$Zh}, \mbox{Zh$\to$En}, and \mbox{En$\to$De} are available at \url{https://huggingface.co/collections/YanZhanPKU/entropy-valley}. The repository documents the required third-party backbones and applicable resource licenses.

\section{Fixed canvas protocol and the length and order dissociation}
\label{sec:appendix_method}

Two protocol assumptions underlie the main text: the target canvas is an external decoding input, and any comparison of length choices needs the reveal order held fixed.

\paragraph{Canvas convention.} Source conditioning is injected via prompt concatenation, so the model input is $[\,\text{prompt}(\xvec)\,] \,\Vert\, \Mask^L$. Decoding runs for $\Tsteps$ steps; each step unmasks a fraction of the canvas under schedule $\pi$. MED reveals the $K_t$ positions with lowest entropy at each step.

\paragraph{Length is external.} LLaDA-style fixed-canvas decoding allocates the target canvas before denoising rather than learning an AR-style stopping rule. The decoder fills the mask slots it is given and has no separate process for choosing how many slots to allocate. Under our shared protocol, $L$ is supplied before denoising and decoded strings are truncated at the first EOS.

\paragraph{Dissociating length from order.} Diffusion MT quality depends on both the chosen canvas length $L$ and the order schedule $\pi$. For order comparisons we fix $L$ to the reference length $\Lref$; for length comparisons we fix $\pi$ to MED at $\Tsteps{=}32$. This separates target canvas choice from reveal order in the comparison reported in \S\ref{sec:analysis-length-order}.

\paragraph{Notation and length unit.} Let $\xvec$ be the source sentence, $\yvec$ the target canvas, and $\Lref$ the reference length. Length is the number of target canvas slots, i.e., the $L$ in $[\,\text{prompt}(\xvec)\,]\Vert\Mask^L$, and excludes the source prompt. The final slot follows the EOS inclusive training format, and the decoded string is truncated at the first EOS. EV excludes this designated EOS slot when averaging entropy, so the score uses the first $L-1$ slots. Candidate ratios are computed against source tokenizer length under the same tokenizer.

\paragraph{Length and order diagnostic.} On \mbox{En$\to$Zh} at $\Tsteps{=}32$ ($N{=}2037$), the static-ratio baseline leaves a $0.0265$ COMET-22 gap to the length oracle. With $L$ held at the reference length, the tested source-guided order schedules span $0.0081$ COMET-22. The EV--Ratio difference is $2.1{\times}$ the tested order span; the reference-length-inclusive span is $3.3{\times}$ the order span. Table~\ref{tab:order_sweep_summary} also reports strict left-to-right and random controls.

\clearpage
\section{Controls for the EV gain}
\label{sec:appendix_ablations}

The controls below examine the candidate range, a global ratio, decode and rerank, decoding budget, probe cost, and sentence-level variation. Subset controls use fixed WMT22 diagnostic subsets with the same prompt, decoding, and evaluation protocol as \S\ref{sec:exp-setup}.

\subsection{Selector checks}
\label{sec:appendix_selector_checks}

These checks use an \mbox{En$\to$Zh} run trained under the same configuration. On a dense sweep over ratios $[0.40,1.50]$ for $500$ sentences, aggregate mean entropy reaches its minimum at ratio $0.70$ ($3.7695$) and rises to $5.3031$ at ratio $1.50$. In the full test set, EV selects all five nominal ratios.

A realized-ratio reassignment preserves the selected ratio distribution while changing the pairing between each source and its canvas. EV remains $+0.0040$ COMET above the sentencewise mean reassignment (95\% CI $[0.0023,0.0058]$). Reassignment within source-length deciles gives the same conclusion. Length-normalized alternatives behave similarly, while unnormalized summed entropy performs worse. These controls support source-conditioned canvas selection and mean entropy as the selection rule.

\subsection{Candidate range sensitivity}

The WMT19 training-corpus median provides a reference scale, while diagnostic range comparisons on WMT22 subsets informed the reported ratio grid. Table~\ref{tab:candidate_range} asks how narrowing or widening that grid changes the result. On these 200-sentence subsets, the En$\to$Zh grid has the lowest length MAE and the highest sacreBLEU, and the En$\to$De grid has the highest subset COMET. The Zh$\to$En subset prefers a wider grid than the reported one, showing sensitivity to higher target length variance.

\begin{table}[H]
\centering
\scriptsize
\setlength{\tabcolsep}{2pt}
\resizebox{\columnwidth}{!}{%
\begin{tabular}{l c c c c c}
\toprule
\textbf{Range} & \textbf{K} & \textbf{Ratios} & \textbf{MAE} & \textbf{sBLEU} & \textbf{COMET} \\
\midrule
\rowcolor{EVGroup}
\multicolumn{6}{l}{\emph{En$\to$Zh (N=200)}} \\
\rowcolor{EVRow}
Reported (narrow) & 5 & [0.70, 0.90] & \textbf{2.71} & \textbf{38.73} & 0.8562 \\
Medium            & 7 & [0.60, 1.00] & 2.84 & 38.49 & \textbf{0.8594} \\
Wide              & 9 & [0.50, 1.10] & 3.06 & 38.17 & 0.8588 \\
Full              & 12 & [0.50, 1.20] & 3.22 & 37.58 & 0.8577 \\
Extra             & 15 & [0.40, 1.50] & 3.89 & 35.98 & 0.8476 \\
\midrule
\rowcolor{EVGroup}
\multicolumn{6}{l}{\emph{En$\to$De (N=200)}} \\
\rowcolor{EVRow}
Reported (narrow) & 5 & [1.50, 1.90] & 4.08 & 22.29 & \textbf{0.7305} (56.0\% gap) \\
Medium            & 7 & [1.30, 2.10] & \textbf{3.90} & \textbf{22.36} & 0.7300 (53.4\%) \\
Wide              & 9 & [1.20, 2.20] & 3.92 & 22.32 & 0.7267 (36.1\%) \\
Full              & 11 & [1.00, 2.40] & 3.98 & 21.97 & 0.7268 (36.6\%) \\
\midrule
\rowcolor{EVGroup}
\multicolumn{6}{l}{\emph{Zh$\to$En (N=200)}} \\
\rowcolor{EVRow}
Reported (narrow)  & 5 & [1.10, 1.30] & --- & 27.64 & 0.8546 \\
Medium            & 7 & [1.05, 1.35] & --- & 27.62 & 0.8554 \\
Wide              & 9 & [1.00, 1.40] & --- & \textbf{27.77} & 0.8566 \\
Full              & 12 & [0.95, 1.50] & --- & 27.63 & \textbf{0.8586} \\
\bottomrule
\end{tabular}%
}
\caption{\textbf{Candidate-range control for Entropy-Valley on fixed 200-sentence WMT22 diagnostic subsets.} The WMT19 training-corpus median provides a reference scale, while diagnostic range comparisons on WMT22 subsets informed the reported ratio grid. The En$\to$Zh grid minimizes length MAE and gives the highest sacreBLEU, while the medium range gives the highest subset COMET; the En$\to$De grid gives the highest subset COMET. The Zh$\to$En subset benefits from a wider range.}
\label{tab:candidate_range}
\end{table}

\subsection{Fixed ratio and multiple candidate controls}

Two simpler alternatives could in principle explain the gain: retune the single global ratio per direction, or decode several neighbouring lengths and rerank. Neither closes the gap. Sweeping post-hoc fixed ratios on the 500-sentence subsets (Table~\ref{tab:ratio_sweep}) leaves EV above the best fixed ratio in all three directions, because per-sentence selection cannot be replaced by one corpus-wide ratio. The heavier alternative in Table~\ref{tab:multi_candidate} decodes five neighbouring canvases to completion and selects by average log-probability; it pays roughly $3.5{\times}$ the cost of Ratio and still scores below EV on the same subset.

\begin{table}[H]
\centering
\footnotesize
\setlength{\tabcolsep}{6pt}
\resizebox{\columnwidth}{!}{%
\begin{tabular}{l c c}
\toprule
\textbf{Method} & \textbf{sacreBLEU} & \textbf{COMET-22} \\
\midrule
\rowcolor{EVGroup}
\multicolumn{3}{l}{\emph{En$\to$Zh (N=500, 32 steps)}} \\
Oracle           & 40.40 & 0.8617 \\
Fixed 0.70       & 31.73 & 0.8114 \\
Fixed 0.75       & 34.59 & 0.8208 \\
Fixed 0.80       & 35.84 & 0.8281 \\
Fixed 0.85       & 36.84 & 0.8354 \\
Fixed 0.90       & 36.93 & 0.8400 \\
\rowcolor{EVRow}
\textbf{Entropy-Valley} & \textbf{38.48} & \textbf{0.8557} \\
\midrule
\rowcolor{EVGroup}
\multicolumn{3}{l}{\emph{En$\to$De (N=500, 32 steps)}} \\
Oracle           & 22.04 & 0.7433 \\
Fixed 1.4        & 16.88 & 0.6930 \\
Fixed 1.5        & 18.57 & 0.7089 \\
Fixed 1.6        & 19.28 & 0.7134 \\
Fixed 1.7        & 20.12 & 0.7187 \\
Fixed 1.8        & 19.64 & 0.7219 \\
Fixed 1.9        & 19.91 & 0.7227 \\
Fixed 2.0        & 18.67 & 0.7186 \\
Fixed 2.1        & 18.05 & 0.7086 \\
Fixed 2.2        & 17.29 & 0.7038 \\
\rowcolor{EVRow}
\textbf{Entropy-Valley} & \textbf{21.24} & \textbf{0.7327} \\
\midrule
\rowcolor{EVGroup}
\multicolumn{3}{l}{\emph{Zh$\to$En (N=500, 32 steps)}} \\
Oracle           & 28.42 & 0.8568 \\
Fixed 1.0        & 19.48 & 0.8099 \\
Fixed 1.1        & 21.71 & 0.8196 \\
Fixed 1.2        & 24.46 & 0.8298 \\
Fixed 1.3        & \textbf{25.57} & 0.8372 \\
Fixed 1.4        & 24.62 & 0.8390 \\
\rowcolor{EVRow}
\textbf{Entropy-Valley} & 25.45 & \textbf{0.8453} \\
\bottomrule
\end{tabular}%
}
\caption{\textbf{Fixed-ratio sweep on fixed 500-sentence WMT22 diagnostic subsets.} EV is compared with fixed ratios around the corpus median for all three directions. On \mbox{Zh$\to$En}, the best post-hoc fixed ratio is $1.4$ at $0.8390$ COMET-22; EV reaches $0.8453$, exceeding it by $+0.0063$ COMET and closing $35.4\%$ of the oracle-length gap on this subset.}
\label{tab:ratio_sweep}
\end{table}

\begin{table}[!htbp]
\centering
\footnotesize
\resizebox{\columnwidth}{!}{%
\begin{tabular}{l c c c}
\toprule
\textbf{Method} & \textbf{sacreBLEU} & \textbf{COMET-22} & \textbf{Cost} \\
\midrule
Oracle length & 40.40 & 0.8617 & $1.00\times$ \\
Fixed ratio 0.8 & 35.84 & 0.8281 & $1.00\times$ \\
\rowcolor{EVRow}
\textbf{EV probe+decode} & \textbf{38.48} & \textbf{0.8557} & $1.15\times$ \\
Decode 5 neighbours & 37.85 & 0.8404 & $\sim 3.5\times$ \\
\bottomrule
\end{tabular}%
}
\caption{\textbf{Multi-candidate length selection on the same En$\to$Zh WMT22 subset used in Table~\ref{tab:ratio_sweep} (N=500, 32 steps).} The alternative decodes five candidate canvas lengths obtained as integer offsets around the EV-selected length, $\{L_{\text{EV}}{-}2, L_{\text{EV}}{-}1, L_{\text{EV}}, L_{\text{EV}}{+}1, L_{\text{EV}}{+}2\}$ (with length collisions deduplicated), to completion and selects by average log-probability; its cost counts all completed decodes.}
\label{tab:multi_candidate}
\end{table}

\subsection{Direct diffusion length baselines}
\label{sec:appendix_direct_baselines}

These controls use one run per direction under the main configuration. Within each direction, every method uses the same model, inputs, and evaluation code. We compare EV with DAEDAL \citep{li2025fixedtrainingfreevariablelengthdenoising} and CAL \citep{liu2026diffusionlmsapproximateoptimal}.

\begin{table*}[!t]
\centering
\scriptsize
\setlength{\tabcolsep}{3pt}
\resizebox{\textwidth}{!}{%
\begin{tabular}{l l c l c c c}
\toprule
\textbf{Dir.} & \textbf{Method} & \textbf{COMET} & \textbf{EV$-$method} & \textbf{Calls} & \textbf{Slots} & \textbf{Latency (s)} \\
\midrule
En$\to$Zh & Ratio & $0.833721$ & $+0.010930$ & $32.00$ & $512.66$ & $0.644$ \\
\rowcolor{EVRow}
En$\to$Zh & \textbf{EV} & $\mathbf{0.844651}$ & -- & $36.09$ & $588.32$ & $0.724$ \\
En$\to$Zh & DAEDAL & $0.828043$ & $+0.016608\;([0.012954,0.020210];\ p<2{\times}10^{-4})$ & $34.58$ & $1175.53$ & $1.798$ \\
En$\to$Zh & CAL & $0.837897$ & $+0.006754\;([0.003726,0.009880];\ p<2{\times}10^{-4})$ & $39.25$ & $628.26$ & $0.773$ \\
\midrule
Zh$\to$En & Ratio & $0.818169$ & $+0.016416$ & $32.00$ & $618.43$ & $0.635$ \\
\rowcolor{EVRow}
Zh$\to$En & \textbf{EV} & $\mathbf{0.834585}$ & -- & $36.43$ & $708.98$ & $0.721$ \\
Zh$\to$En & DAEDAL & $0.812488$ & $+0.022097\;([0.018849,0.025375];\ p<2{\times}10^{-4})$ & $37.49$ & $1533.16$ & $1.972$ \\
Zh$\to$En & CAL & $0.833352$ & $+0.001233\;([-0.001248,0.003686];\ p=0.3342)$ & $39.42$ & $786.51$ & $0.770$ \\
\bottomrule
\end{tabular}%
}
\caption{\textbf{Direct diffusion length comparisons on WMT22 ($N{=}2037$ per direction).} Confidence intervals and paired bootstrap $p$-values are shown for DAEDAL and CAL. Calls and Slots are per-sentence averages; latency is measured in seconds per sentence in the reported setup. EV is above DAEDAL in both directions and CAL on En$\to$Zh. The EV and CAL interval on Zh$\to$En includes zero. EV uses fewer calls and slots than CAL, with lower latency in this setup.}
\label{tab:direct_length_baselines}
\end{table*}

\FloatBarrier

\subsection{Decoding budget robustness}
\label{sec:appendix_ablations_pareto}

The next control varies the decoding-step budget $T \in \{2,4,8,16,32,64,128\}$ on the full WMT22 test set for one analyzed run ($N{=}2037$). On En$\to$Zh, EV is at least as high as Ratio at every step budget. On En$\to$De, EV leads at small $T$ and falls slightly below Ratio at $T{\ge}64$; this is the boundary recorded in the Limitations section.

\begin{figure*}[t]
  \centering
  \includegraphics[width=0.94\textwidth]{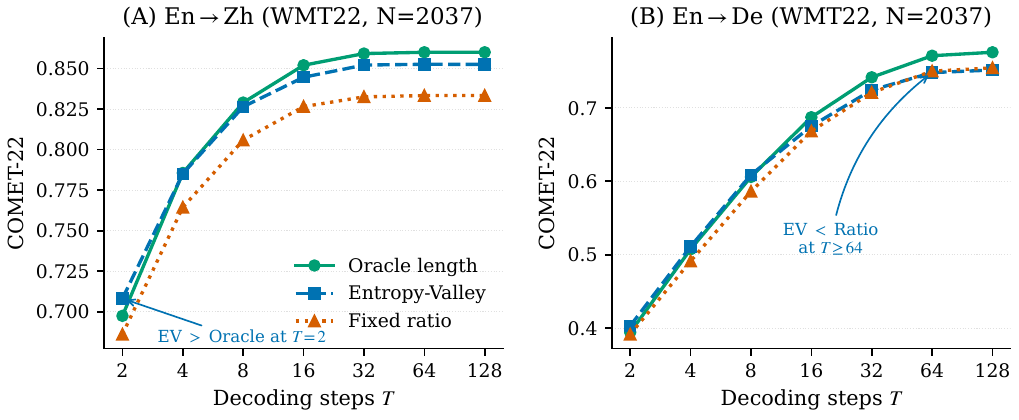}
  \caption{\textbf{Cost--quality Pareto.}
COMET-22 vs decoding steps $T \in \{2,4,8,16,32,64,128\}$ on WMT22 ($N{=}2037$, one analyzed run).
(A) En$\to$Zh: EV is at least as high as Ratio at every $T$.
(B) En$\to$De: EV is higher at small $T$ and slightly lower than Ratio at $T \geq 64$.
Oracle is a length oracle, not a decode oracle.}
  \label{fig:pareto}
\end{figure*}

\begin{table*}[!t]
\centering
\footnotesize
\setlength{\tabcolsep}{5pt}
\resizebox{\textwidth}{!}{%
\begin{tabular}{l cccc cccc}
\toprule
 & \multicolumn{4}{c}{\textbf{En$\to$Zh}} & \multicolumn{4}{c}{\textbf{En$\to$De}} \\
 \cmidrule(lr){2-5}\cmidrule(lr){6-9}
\textbf{Steps} & Oracle & EV & Ratio & Gap\% & Oracle & EV & Ratio & Gap\% \\
\midrule
2     & 0.6975 & 0.7082 & 0.6861 & \textbf{194\%} & 0.3942 & 0.4030 & 0.3915 & \textbf{426\%} \\
4     & 0.7856 & 0.7851 & 0.7644 & 98\%           & 0.5074 & 0.5112 & 0.4912 & \textbf{123\%} \\
8     & 0.8290 & 0.8264 & 0.8058 & 89\%           & 0.6060 & 0.6089 & 0.5862 & \textbf{115\%} \\
16    & 0.8520 & 0.8446 & 0.8265 & 71\%           & 0.6873 & 0.6752 & 0.6683 & 36\% \\
32    & 0.8592 & 0.8521 & 0.8325 & 73\%           & 0.7417 & 0.7243 & 0.7209 & 16\% \\
64    & 0.8600 & 0.8526 & 0.8333 & 72\%           & 0.7711 & 0.7481 & 0.7501 & $-10\%$ \\
128   & 0.8600 & 0.8525 & 0.8333 & 72\%           & 0.7758 & 0.7516 & 0.7545 & $-14\%$ \\
\bottomrule
\end{tabular}%
}
\caption{\textbf{Full cost--quality Pareto on WMT22 ($N{=}2037$, one analyzed run).} Gap\% is computed relative to the Ratio and Oracle length columns at the same step budget; values can exceed 100\% because Oracle fixes reference length but does not optimize decoding under small $T$.}
\label{tab:full_pareto}
\end{table*}

\FloatBarrier

\subsection{Paired significance and compute budget}

The final controls address statistical noise and probe cost. Table~\ref{tab:full_significance} reports paired bootstrap and Wilcoxon tests for the En$\leftrightarrow$Zh and En$\to$De main comparisons. The En$\to$Zh comparison is reliable in run A, and Zh$\to$En is reliable across all three runs. En$\to$De passes Wilcoxon but not bootstrap, matching the weaker sentence-level evidence noted in \S\ref{sec:exp-main}. Table~\ref{tab:fwdcount_control} evaluates Ratio at EV's maximum total forward pass budget, $T{=}37$, and at a larger $T{=}40$ budget. The extra budget recovers at most $0.001$ COMET, well below the En$\leftrightarrow$Zh EV gains, so the gain is the length choice rather than the additional compute.

\begin{table}[H]
\centering
\scriptsize
\setlength{\tabcolsep}{2pt}
\resizebox{\columnwidth}{!}{%
\begin{tabular}{l l c c c c}
\toprule
\textbf{Dir.} & \textbf{Comp.} & $\Delta$ & 95\% CI & Boot. $p$ & Wilc. $p$ \\
\midrule
En$\to$Zh (A) & EV--Ratio      & $+0.0196$ & $[+0.0166,+0.0226]$ & $<10^{-4}$ & $4.1{\times}10^{-32}$ \\
En$\to$Zh (A) & Oracle--EV     & $+0.0071$ & $[+0.0046,+0.0096]$ & $<10^{-4}$ & $1.3{\times}10^{-14}$ \\
En$\to$Zh (A) & Oracle--Ratio  & $+0.0266$ & $[+0.0233,+0.0300]$ & $<10^{-4}$ & $3.4{\times}10^{-58}$ \\
\midrule
En$\to$De (A) & EV--Ratio      & $+0.0034$ & $[-0.0016,+0.0086]$ & $0.091$ & $8.3{\times}10^{-3}$ \\
En$\to$De (A) & Oracle--EV     & $+0.0173$ & $[+0.0121,+0.0225]$ & $<10^{-4}$ & $1.6{\times}10^{-9}$ \\
En$\to$De (A) & Oracle--Ratio  & $+0.0208$ & $[+0.0155,+0.0261]$ & $<10^{-4}$ & $6.0{\times}10^{-14}$ \\
\midrule
Zh$\to$En (A) & EV--Ratio & $+0.0174$ & $[+0.0150,+0.0198]$ & $<10^{-4}$ & $6.9{\times}10^{-45}$ \\
Zh$\to$En (B) & EV--Ratio & $+0.0154$ & $[+0.0131,+0.0178]$ & $<10^{-4}$ & $1.9{\times}10^{-41}$ \\
Zh$\to$En (C) & EV--Ratio & $+0.0167$ & $[+0.0143,+0.0192]$ & $<10^{-4}$ & $2.1{\times}10^{-45}$ \\
\bottomrule
\end{tabular}%
}
\caption{\textbf{Sentence-level paired COMET-22 tests for LLaDA on WMT22 ($N{=}2037$ throughout).} Positive $\Delta$ means the first method is better. Paired bootstrap uses 10k resamples; Wilcoxon uses the signed-rank test with Pratt handling of zeros. The table reports run A for \mbox{En$\to$Zh} and \mbox{En$\to$De}, plus all three \mbox{Zh$\to$En} runs. Oracle rows show the run A gap to the reference-length upper bound.}
\label{tab:full_significance}
\end{table}

\begin{table}[H]
\centering
\footnotesize
\setlength{\tabcolsep}{4pt}
\resizebox{\columnwidth}{!}{%
\begin{tabular}{l l c c r c c}
\toprule
\textbf{Direction} & \textbf{Method} & \textbf{T} & \textbf{Probe cap} & \textbf{Fwd cap} & \textbf{COMET-22} & \textbf{sacreBLEU} \\
\midrule
\rowcolor{EVGroup}
\multicolumn{7}{l}{\emph{En$\to$Zh, one analyzed run, WMT22 N=2037}} \\
En$\to$Zh & Ratio 0.8       & 32 & 0 & 32 & 0.8325 & 36.53 \\
\rowcolor{EVRow}
En$\to$Zh & \textbf{\method} & \textbf{32} & $\mathbf{\leq 5}$ & $\mathbf{\leq 37}$ & $\mathbf{0.8521}$ & $\mathbf{38.56}$ \\
En$\to$Zh & Ratio 0.8       & 37 & 0 & 37 & 0.8334 & 36.66 \\
En$\to$Zh & Ratio 0.8       & 40 & 0 & 40 & 0.8333 & 36.71 \\
\midrule
\rowcolor{EVGroup}
\multicolumn{7}{l}{\emph{Zh$\to$En, one analyzed run, WMT22 N=2037}} \\
Zh$\to$En & Ratio 1.2       & 32 & 0 & 32 & 0.8260 & 23.53 \\
\rowcolor{EVRow}
Zh$\to$En & \textbf{\method} & \textbf{32} & $\mathbf{\leq 5}$ & $\mathbf{\leq 37}$ & $\mathbf{0.8434}$ & $\mathbf{25.16}$ \\
Zh$\to$En & Ratio 1.2       & 37 & 0 & 37 & 0.8266 & 23.71 \\
Zh$\to$En & Ratio 1.2       & 40 & 0 & 40 & 0.8269 & 23.73 \\
\bottomrule
\end{tabular}%
}
\caption{\textbf{Forward-pass budget control.} Ratio is evaluated at the default budget ($T{=}32$), EV's maximum total forward pass budget ($T{=}37$), and a larger budget ($T{=}40$). Duplicate integer candidate lengths can make EV's realized budget smaller than the cap.}
\label{tab:fwdcount_control}
\end{table}

\FloatBarrier

\section{Coverage losses from short canvases}
\label{sec:appendix_coverage}

The analysis in this section shows that EV preserves source-side content that short Ratio canvases can omit. The evidence differs across directions and length buckets.

\paragraph{Error-type categories.} For every WMT22 source sentence we extract three categories whose retention can be detected by literal substring match in the target output: anonymisation \textbf{placeholders} (\texttt{\#PRS\_ORG\#}, \texttt{\#DATE\#}, etc.), \textbf{numbers} (integers, decimals, percentages, dollar amounts), and \textbf{capitalised named-entity tokens} (heuristic; \mbox{En$\to$Zh} only, since Chinese sources do not use capitalisation). Each sentence is also bucketed by the LLaDA-token-level reference-to-source ratio $r=|\Lref|/|\xvec|$. Retention rates are averaged per sentence over the three runs used in \S\ref{sec:exp-main}, and the EV$-$Ratio difference is reported with paired bootstrap 95\% CIs ($10$k resamples). The outputs are not re-decoded for this analysis.

\paragraph{Literal retention.} On \mbox{En$\to$Zh}, placeholder retention rises from $66.9\%$ to $89.1\%$ ($+22.1$ pp) and number retention from $77.6\%$ to $81.3\%$ ($+3.7$ pp), both with bootstrap CIs strictly above zero (Table~\ref{tab:coverage}). Capitalised-NE retention moves little because \mbox{En$\to$Zh} often transliterates entities; this category is heuristic rather than diagnostic. On \mbox{Zh$\to$En}, Ratio already retains $96.6\%$ of placeholders and $80.7\%$ of numbers, so the EV gains of $+2.9$ and $+1.9$ pp are too small to account for the \mbox{Zh$\to$En} COMET improvement on their own. The \mbox{Zh$\to$En} mechanism must involve non-literal content that our error-type categories do not directly probe.

\paragraph{Length buckets.} The largest \mbox{En$\to$Zh} COMET gain ($+0.0346$) is in the dominant $r{\ge}0.8$ bucket where the fixed candidate grid is well aligned. The $r{<}0.6$ bucket still improves on average ($+0.0054$), but individual sentences in this regime can fail when the needed compression lies below the grid; \S\ref{sec:appendix_cases} gives such a case.

\begin{table*}[!t]
\centering
\footnotesize
\setlength{\tabcolsep}{4pt}
\resizebox{\textwidth}{!}{%
\begin{tabular}{l r r r r l}
\toprule
\textbf{Category} & \textbf{N} & \textbf{Ratio} & \textbf{EV} & \textbf{$\Delta$ (pp)} & \textbf{95\% CI (pp)} \\
\midrule
\rowcolor{EVGroup}
\multicolumn{6}{l}{\emph{En$\to$Zh --- error-type preservation rate}} \\
Placeholder (\#PRS\_ORG\#) & 58 & 66.9\% & 89.1\% & +22.1$^{*}$ & [+12.4, +32.5] \\
Number / percentage / decimal & 290 & 77.6\% & 81.3\% & +3.7$^{*}$ & [+1.4, +6.1] \\
Capitalized named entity (heuristic) & 879 & 11.2\% & 11.9\% & +0.7$^{*}$ & [+0.0, +1.4] \\
\midrule
\rowcolor{EVGroup}
\multicolumn{6}{l}{\emph{En$\to$Zh --- length-compression buckets (LLaDA token ratio)}} \\
\textbf{Bucket} & \textbf{N} & \textbf{Ratio COMET} & \textbf{EV COMET} & \textbf{$\Delta$ COMET} & \textbf{95\% CI} \\
$r<0.6$ & 236 & 0.8449 & 0.8503 & $+0.0054^{*}$ & $[+0.0016, +0.0093]$ \\
$0.6\le r<0.8$ & 796 & 0.8630 & 0.8678 & $+0.0048^{*}$ & $[+0.0016, +0.0081]$ \\
$r\ge 0.8$ & 1005 & 0.8056 & 0.8402 & $+0.0346^{*}$ & $[+0.0295, +0.0401]$ \\
\midrule
\rowcolor{EVGroup}
\multicolumn{6}{l}{\emph{Zh$\to$En --- error-type preservation rate}} \\
Placeholder (\#PRS\_ORG\#) & 58 & 96.6\% & 99.4\% & +2.9 & [-1.1, +8.6] \\
Number / percentage / decimal & 303 & 80.7\% & 82.6\% & +1.9 & [0.0, +4.0] \\
\bottomrule
\end{tabular}%
}
\caption{\textbf{Coverage and error categories.} Top: source-side literal retention rate, averaged over three runs. Middle: En$\to$Zh COMET-22 grouped by $r=|\Lref|/|\xvec|$, the reference target length divided by source length under the LLaDA tokenizer. Bottom: Zh$\to$En literal retention. $\Delta$ is in percentage points for retention rows; $^{*}$ marks an interval whose unrounded lower endpoint is above zero.}
\label{tab:coverage}
\end{table*}

The middle block of Table~\ref{tab:coverage} gives the bucket values used for the fixed-grid boundary discussion in \S\ref{sec:limitations-candidate-grid}.

\section{Fixed grid wins and boundary failures}
\label{sec:appendix_cases}

The cases that follow turn Figure~\ref{fig:cases} into one win, one win, and one loss for the fixed candidate grid. Each case lists the source, reference length, the canvases chosen by Ratio and EV, and the resulting COMET. The Chinese outputs are rendered as raster images to avoid font issues.

\noindent \textbf{Coverage.} Source: \textit{``Tap Reset Now.''}; reference length $\Lref{=}6$. \ratio's $L{=}5$ canvas collapses to a single time adverb (COMET $0.47$); \method's $L{=}6$ canvas recovers the action verb (COMET $\mathbf{0.91}$, $\Delta{=}{+}0.44$).

\noindent \textbf{Placeholder preservation.} Source: \textit{``Under \#PRS\_ORG\#, tap Sign out.''}; reference length $\Lref{=}12$. \ratio's $L{=}10$ canvas drops the \#PRS\_ORG\# placeholder (COMET $0.43$); \method's $L{=}12$ canvas preserves it verbatim (COMET $\mathbf{0.87}$, $\Delta{=}{+}0.43$).

\noindent \textbf{Candidate-grid boundary.} Source: \textit{``Please give me a moment.''}; reference length $\Lref{=}6$, true compression ratio $r{\approx}0.6$. \ratio's $L{=}6$ canvas saturates (COMET $\mathbf{0.93}$); \method's $L{=}7$ canvas overshoots (COMET $0.77$). The needed canvas lies below the fixed grid $[0.70,0.90]$, so the entropy score cannot reach it.

\paragraph{Aggregate losses.}
The three cases generalise to the full failure population in Table~\ref{tab:failure_buckets}. In the analyzed run, EV scores below Ratio on 596 \mbox{En$\to$Zh} sentences; the dominant cause is a within-grid error, where EV picks a neighbouring canvas although a better canvas is represented in the grid. Most losses are selection errors within the fixed grid rather than grid-boundary failures.

\begin{table}[!htbp]
\centering
\footnotesize
\setlength{\tabcolsep}{3pt}
\begin{tabular}{l r r r}
\toprule
\textbf{Likely cause} & \textbf{N} & \textbf{Share} & \textbf{Mean $\Delta$COMET} \\
\midrule
Overshoot beyond grid & 129 & 21.6\% & $-0.0278$ \\
Under-compression & 126 & 21.1\% & $-0.0206$ \\
Within-grid wrong pick & 339 & 56.9\% & $-0.0247$ \\
Very short source & 2 & 0.3\% & --- \\
\bottomrule
\end{tabular}
\caption{\textbf{Failure buckets for the 596 \mbox{En$\to$Zh} cases where EV trails Ratio in one analyzed run (WMT22, $N{=}2037$).} Buckets are assigned by mutually exclusive rules.}
\label{tab:failure_buckets}
\end{table}

\section{Reveal order controls and negative diagnostics}
\label{sec:appendix_negatives}

These diagnostics test whether the evaluated reveal-order schedules or a chat-aligned masked-diffusion backbone close the Ratio--Oracle gap under the same fixed-canvas protocol. Neither does so in the tested settings.

\subsection{Source-guided order schedules under matched length}
\label{sec:appendix_negatives_sig}

Holding $L$ at the reference length removes length choice from the comparison, so any remaining variation comes from the reveal order. Under this protocol at $\Tsteps{=}32$ on full WMT22 \mbox{En$\to$Zh} ($N{=}2037$), six source-guided variants span $0.0081$ COMET, and none reliably improves over MED. Strict left-to-right and random order are additional controls, and both score below MED. The result covers the evaluated schedules rather than all reveal policies.

\begin{table}[!htbp]
\centering
\footnotesize
\setlength{\tabcolsep}{4pt}
\resizebox{\columnwidth}{!}{%
\begin{tabular}{l c c c}
\toprule
\textbf{Schedule} & \textbf{COMET-22} & \textbf{$\Delta$ vs MED} & \textbf{Paired bootstrap $p$} \\
\midrule
MED (baseline)         & 0.8597 & ---       & --- \\
SIG-first              & 0.8603 & $+0.0006$ & $0.62$ \\
Reverse-SIG            & 0.8523 & $-0.0075$ & $<10^{-4}$ \\
Hybrid MED$\to$SIG, $p{=}0.25$ & 0.8582 & $-0.0015$ & $0.21$ \\
Hybrid MED$\to$SIG, $p{=}0.50$ & 0.8575 & $-0.0022$ & $0.011$ \\
Hybrid MED$\to$SIG, $p{=}0.75$ & 0.8591 & $-0.0007$ & $0.28$ \\
Entropy-weighted SIG   & 0.8591 & $-0.0006$ & $0.38$ \\
\addlinespace
\rowcolor{EVGroup}
\multicolumn{4}{l}{\emph{Additional reveal-order controls}} \\
Strict L2R              & 0.8510 & $-0.0088$ & -- \\
Random                  & 0.8539 & $-0.0059$ & $<10^{-4}$ \\
\midrule
\textbf{Source-guided span} (max$-$min) & \textbf{0.0081} & --- & --- \\
\bottomrule
\end{tabular}
}
\caption{\textbf{Reveal-order controls on WMT22 En$\to$Zh ($N{=}2037$, $\Tsteps{=}32$) with reference length and the tokens-per-step schedule fixed.} The six source-guided variants span $0.0081$ COMET. Strict left-to-right and random order are additional controls and both score below MED. A paired bootstrap value is not reported for Strict L2R.}
\label{tab:order_sweep_summary}
\end{table}

SIG-first might still help on subsets where source-guided unmasking has the strongest case, such as dates, enumerations, named entities, or numbers. On a 400-sentence challenge subset stratified by these categories, SIG-first averages $-0.0004$ COMET relative to MED (Table~\ref{tab:negative_order}), and a sentence-level SIG-Concentration score does not identify a useful subset either (Spearman $\rho = -0.0675$ with SIG-first benefit, $p = 0.13$). On this subset MED is again at least as good as the alternatives, so the negative result for the tested schedules holds at the category level as well.

\begin{table}[!htbp]
\centering
\footnotesize
\resizebox{\columnwidth}{!}{%
\begin{tabular}{l r c c c}
\toprule
\textbf{Category} & \textbf{N} & \textbf{SIG-first} & \textbf{MED} & \textbf{$\Delta$ (SIG$-$MED)} \\
\midrule
Dates           &  32 & 0.8652 & 0.8665 & $-0.0013$ \\
Enumeration     &  75 & 0.8560 & 0.8533 & $+0.0027$ \\
Named entities  & 280 & 0.8627 & 0.8629 & $-0.0002$ \\
Numbers         & 123 & 0.8679 & 0.8724 & $-0.0045$ \\
\midrule
\textbf{Overall} & \textbf{400} & $\mathbf{0.8640}$ & $\mathbf{0.8644}$ & $\mathbf{-0.0004}$ \\
\bottomrule
\end{tabular}%
}
\caption{\textbf{Source-guided order diagnostic on an En$\to$Zh challenge subset ($N{=}400$, one analyzed run, $\Tsteps{=}32$ MED vs.\ SIG-first under matched oracle length).} SIG-first reveals high-source-dependence slots first; MED is the confidence-first schedule used in the main experiments.}
\label{tab:negative_order}
\end{table}

\subsection{Dream-Instruct protocol mismatch}
\label{sec:appendix_dream7b}

We also tested \textsc{Dream-v0-Instruct-7B}. Under the LLaDA-compatible MED fixed-canvas protocol used everywhere else in this paper, it does not produce usable MT outputs, which is why the cross-backbone check in \S\ref{sec:appendix_crossbb} uses \textsc{Dream-v0-Base-7B} instead. We tested three configurations:
\begin{enumerate}[leftmargin=*,topsep=2pt,itemsep=1pt]
  \item \textbf{LoRA-SFT with plain prompts.} LoRA (rank 64) on 200k En$\to$Zh pairs using the LLaDA-compatible plain-text template (9{,}660 logged steps). Loss plateaued near the random baseline $\log|V|\approx 11.93$ (Qwen2.5 vocabulary $|V|{=}152{,}064$): final loss $12.65$ and last-10-step mean $12.77 \pm 0.88$, with the mean over steps $\geq 1000$ at $13.22$. Under this LoRA setup, training does not override Dream-Instruct's chat-aligned predictive distribution.
  \item \textbf{LoRA-SFT with Qwen chat template.} Training loss reaches a minimum of $11.9162$ at step 2{,}120, indistinguishable from the uniform-distribution baseline $\log|V|\approx 11.93$; the decoder still emits \texttt{<|im\_end|>} at the canvas tail and most outputs are truncated to the empty string.
  \item \textbf{Zero-shot MED decoding.} Without any SFT, EOS confidence at the canvas tail is uniformly high across candidate lengths, so the entropy signal is uninformative under this protocol. On WMT22 \mbox{En$\to$Zh} (N=500), the empty-output rate is $97.4\%$ (EV), $96.6\%$ (Ratio), and $94.8\%$ (Oracle); non-empty outputs are short repetition patterns (e.g.~``\textit{....}'', ``\textit{992255}'') with mean length $0.17$--$0.97$ characters versus a $\sim 25$-character reference.
\end{enumerate}
The shared failure mode is a chat-aligned EOS at the canvas tail that no length choice can repair. Dream-Base and DiffuLLaMA do not have this failure mode under the same MED protocol, which is why the cross-backbone scope check is run there.

\section{Scope across masked diffusion backbones}
\label{sec:appendix_crossbb}

The cross-backbone check asks whether the gain reported on LLaDA is specific to that backbone. We re-run the \oracleref, \ratio, and \method protocol of \S\ref{sec:exp-main} on two other pretrained masked-diffusion models, \textsc{Dream-v0-Base-7B} \citep{ye2025dream}, a Qwen2.5-7B initialization continued-pretrained on $580$B tokens with a masked-diffusion objective, and \textsc{DiffuLLaMA-7B} \citep{gong2025diffullama}, a Llama-2-7B initialization continued-pretrained with the MDLM absorbing-diffusion objective. The protocol is held fixed: $200$k SFT pairs per direction, LoRA ($r{=}64$, $\alpha{=}128$, dropout $0.05$; per-backbone $7$-module target set), $\Tsteps{=}32$ MED decoding, and three runs per direction. We use Dream-Base rather than Dream-Instruct because the latter is not usable under this protocol (\S\ref{sec:appendix_dream7b}).

\begin{table*}[!htbp]
\centering
\small
\setlength{\tabcolsep}{1.8pt}
\begin{tabular}{l l c c c c c}
\toprule
\textbf{Backbone} & \textbf{Direction} & \textbf{Oracle COMET} & \textbf{Ratio COMET} & \textbf{EV COMET} & \textbf{$\Delta$ EV$-$Ratio} & \textbf{COMET Gap closure} \\
\midrule
\rowcolor{EVGroup}
\multicolumn{7}{l}{\textit{LLaDA-8B-Base (paper main results, \S\ref{sec:exp-main}; reproduced for cross-backbone comparison)}} \\
LLaDA-8B-Base   & En$\to$Zh & $0.8610 {\pm} 0.0013$ & $0.8345 {\pm} 0.0017$ & \cellcolor{EVCell}$\mathbf{0.8517 {\pm} 0.0006}$ & $+0.0172$ & \cellcolor{EVCell}$\mathbf{64.9 {\pm} 7.4\%}$ \\
LLaDA-8B-Base   & Zh$\to$En & $0.8519 {\pm} 0.0007$ & $0.8266 {\pm} 0.0010$ & \cellcolor{EVCell}$\mathbf{0.8431 {\pm} 0.0004}$ & $+0.0165$ & \cellcolor{EVCell}$\mathbf{65.3 {\pm} 0.8\%}$ \\
LLaDA-8B-Base   & En$\to$De & $0.7382 {\pm} 0.0090$ & $0.7170 {\pm} 0.0090$ & \cellcolor{EVCell}$\mathbf{0.7240 {\pm} 0.0078}$ & $+0.0070$ & \cellcolor{EVCell}$\mathbf{33.0 {\pm} 8.4\%}$ \\
\midrule
\rowcolor{EVGroup}
\multicolumn{7}{l}{\textit{Dream-v0-Base-7B (Qwen2.5-7B continued-pretrained; Qwen2 tokenizer, $152$k vocab)}} \\
Dream-Base      & En$\to$Zh & $0.8414 {\pm} 0.0020$ & $0.8002 {\pm} 0.0016$ & \cellcolor{EVCell}$\mathbf{0.8239 {\pm} 0.0004}$ & $+0.0238$ & \cellcolor{EVCell}$\mathbf{57.7 {\pm} 3.1\%}$ \\
Dream-Base      & Zh$\to$En & $0.8490 {\pm} 0.0008$ & $0.8303 {\pm} 0.0009$ & \cellcolor{EVCell}$\mathbf{0.8431 {\pm} 0.0009}$ & $+0.0128$ & \cellcolor{EVCell}$\mathbf{68.7 {\pm} 6.2\%}$ \\
Dream-Base      & En$\to$De & $0.7271 {\pm} 0.0017$ & $0.6904 {\pm} 0.0028$ & \cellcolor{EVCell}$\mathbf{0.7106 {\pm} 0.0026}$ & $+0.0201$ & \cellcolor{EVCell}$\mathbf{54.4 {\pm} 8.8\%}$ \\
\midrule
\rowcolor{EVGroup}
\multicolumn{7}{l}{\textit{DiffuLLaMA-7B (Llama-2-7B continued-pretrained; Llama-2 tokenizer, $32$k vocab)}} \\
DiffuLLaMA      & En$\to$Zh & $0.7390 {\pm} 0.0041$ & $0.6010 {\pm} 0.0021$ & \cellcolor{EVCell}$\mathbf{0.6172 {\pm} 0.0035}$ & $+0.0162$ & \cellcolor{EVCell}$11.8 {\pm} 3.5\%$ \\
DiffuLLaMA      & Zh$\to$En & $0.8261 {\pm} 0.0016$ & $0.6612 {\pm} 0.0040$ & \cellcolor{EVCell}$\mathbf{0.6980 {\pm} 0.0026}$ & $+0.0368$ & \cellcolor{EVCell}$22.3 {\pm} 2.6\%$ \\
DiffuLLaMA      & En$\to$De & $0.6909 {\pm} 0.0369$ & $0.6439 {\pm} 0.0348$ & \cellcolor{EVCell}$\mathbf{0.6749 {\pm} 0.0306}$ & $+0.0310$ & \cellcolor{EVCell}$\mathbf{66.1 {\pm} 13.0\%}$ \\
\bottomrule
\end{tabular}
\caption{\textbf{Cross-backbone COMET-22 results for \method (WMT22, $N{=}2037$, three runs per aggregate cell, $\Tsteps{=}32$ MED).} Values are mean $\pm$ std across runs. COMET-22 gap closure is $(\text{EV}-\text{Ratio})/(\text{Oracle}-\text{Ratio})$ per run, then mean $\pm$ std. EV is above Ratio in all six Dream-Base and DiffuLLaMA aggregate COMET cells; paired tests for every run give $p<10^{-3}$ in all $18$ Dream-Base and DiffuLLaMA comparisons, with positive 95\% bootstrap confidence intervals. Absolute scores and closures vary by backbone.}
\label{tab:crossbb}
\end{table*}

\paragraph{Interpretation across backbones.}
EV remains above Ratio in every Dream-Base and DiffuLLaMA cell of Table~\ref{tab:crossbb}, with the same pattern in the sacreBLEU companion view (Table~\ref{tab:crossbb_bleu}). The size of the gain, and the share of the Ratio--Oracle gap it closes, change with the backbone: Dream-Base closes between $54\%$ and $69\%$ across directions; DiffuLLaMA closes more on En$\to$De ($66.1\%$) than on the Chinese-side directions, consistent with its $32$k Llama-2 tokenizer covering Chinese less well than the $152$k Qwen2 tokenizer used by Dream-Base. These runs change tokenizer, initialization, pretraining mixture, and architecture together, so they check scope but do not attribute the variation to a single factor.

\begin{table}[!htbp]
\centering
\scriptsize
\setlength{\tabcolsep}{1.8pt}
\begin{tabular}{l l c c c}
\toprule
\textbf{Backbone} & \textbf{Direction} & \textbf{Oracle} & \textbf{Ratio} & \textbf{EV} \\
\midrule
LLaDA-8B-Base  & En$\to$Zh & $40.81 {\pm} 0.23$ & $36.72 {\pm} 0.22$ & \cellcolor{EVCell}$\mathbf{38.57 {\pm} 0.13}$ \\
LLaDA-8B-Base  & Zh$\to$En & $27.93 {\pm} 0.23$ & $23.65 {\pm} 0.21$ & \cellcolor{EVCell}$\mathbf{25.28 {\pm} 0.33}$ \\
LLaDA-8B-Base  & En$\to$De & $22.55 {\pm} 0.77$ & $20.73 {\pm} 0.68$ & \cellcolor{EVCell}$\mathbf{21.55 {\pm} 0.85}$ \\
\midrule
Dream-Base     & En$\to$Zh & $36.30 {\pm} 0.22$ & $29.37 {\pm} 0.20$ & \cellcolor{EVCell}$\mathbf{32.83 {\pm} 0.28}$ \\
Dream-Base     & Zh$\to$En & $27.24 {\pm} 0.28$ & $23.56 {\pm} 0.04$ & \cellcolor{EVCell}$\mathbf{25.78 {\pm} 0.20}$ \\
Dream-Base     & En$\to$De & $21.64 {\pm} 0.16$ & $17.31 {\pm} 0.25$ & \cellcolor{EVCell}$\mathbf{20.06 {\pm} 0.17}$ \\
\midrule
DiffuLLaMA     & En$\to$Zh & $24.91 {\pm} 0.16$ & $\phantom{0}7.49 {\pm} 0.12$ & \cellcolor{EVCell}$\mathbf{\phantom{0}7.92 {\pm} 0.10}$ \\
DiffuLLaMA     & Zh$\to$En & $22.63 {\pm} 0.30$ & $\phantom{0}9.36 {\pm} 0.32$ & \cellcolor{EVCell}$\mathbf{10.80 {\pm} 0.14}$ \\
DiffuLLaMA     & En$\to$De & $19.72 {\pm} 2.64$ & $14.95 {\pm} 1.89$ & \cellcolor{EVCell}$\mathbf{17.94 {\pm} 1.99}$ \\
\bottomrule
\end{tabular}
\caption{\textbf{Companion sacreBLEU view of Table~\ref{tab:crossbb} (WMT22, $N{=}2037$, three runs per cell).} Values are mean $\pm$ std across runs.}
\label{tab:crossbb_bleu}
\end{table}

\section{Additional translation direction}
\label{sec:appendix_defr}

We evaluate De$\to$Fr to add a direction with no English on either side. Two nested WMT19 training pools use the same $500$ update budget and candidate set for the direction. Ratio is $1.0$, and the fixed EV candidate set is $\{0.8,0.9,1.0,1.1,1.2\}$. The trained systems and candidate set are evaluated on a WMT19 held out set and transferred unchanged to FLORES devtest.

\begin{table}[!htbp]
\centering
\scriptsize
\setlength{\tabcolsep}{2pt}
\resizebox{\columnwidth}{!}{%
\begin{tabular}{l c c c c}
\toprule
\textbf{Evaluation set} & \textbf{Pool} & \textbf{Ratio} & \textbf{EV} & \textbf{$\Delta$COMET (95\% CI)} \\
\midrule
WMT19 held out & 50k & $0.7428$ & \cellcolor{EVCell}$0.7514$ & $+0.0086\ [0.0034,0.0139]$ \\
WMT19 held out & 200k & $0.7386$ & \cellcolor{EVCell}$0.7485$ & $+0.0100\ [0.0050,0.0152]$ \\
FLORES devtest & 50k & $0.7398$ & \cellcolor{EVCell}$0.7481$ & $+0.0082\ [0.0020,0.0143]$ \\
FLORES devtest & 200k & $0.7315$ & \cellcolor{EVCell}$0.7403$ & $+0.0088\ [0.0024,0.0152]$ \\
\bottomrule
\end{tabular}%
}
\caption{\textbf{De$\to$Fr results for two nested training pools.} WMT19 held out contains $N{=}1512$ sentences, and FLORES uses the full $N{=}1012$ De$\to$Fr devtest. Both pools use a fixed $500$ update budget and one candidate set for the direction. EV exceeds Ratio in all four settings, with positive paired confidence intervals; paired bootstrap and Wilcoxon tests are below $0.05$ throughout. This is a robustness check under a fixed update budget, not a comparison of scaling with data size.}
\label{tab:defr_transfer}
\end{table}

\section{Bilingual human evaluation protocol}
\label{sec:appendix_human_protocol}

\begin{figure*}[!t]
  \centering
  \includegraphics[width=\textwidth]{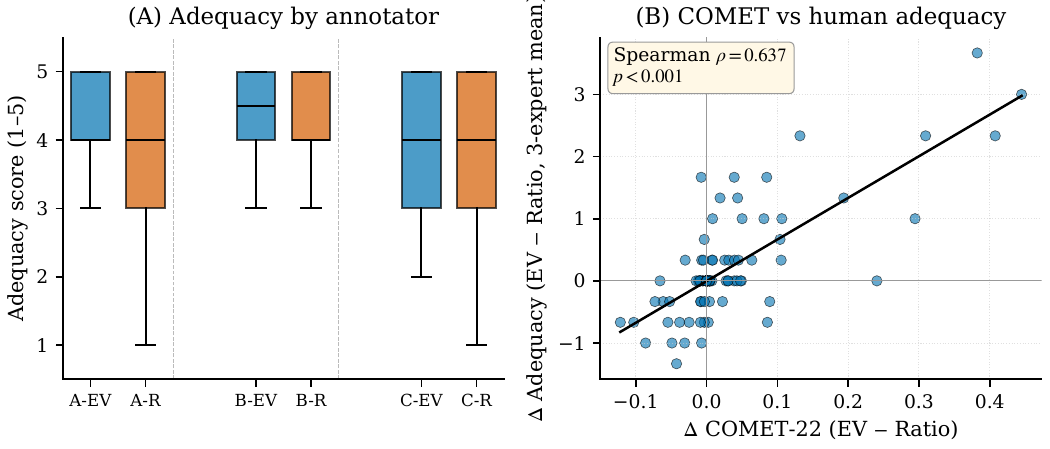}
  \caption{\textbf{Human evaluation on En$\to$Zh.} (A) Per-annotator adequacy boxplots. (B) Sentence-level $\Delta$COMET-22 vs 3-expert mean $\Delta$Adequacy (EV $-$ Ratio); Spearman $\rho=0.637$ ($p<0.001$). The corresponding \mbox{Zh$\to$En} correlation is reported in Table~\ref{tab:human_eval_details}.}
  \label{fig:human_eval}
\end{figure*}

The protocol below supports the \mbox{En$\leftrightarrow$Zh} human study in Table~\ref{tab:human_eval} and the per-annotator breakdown in Figure~\ref{fig:human_eval} and Table~\ref{tab:human_eval_details}. The same three professional translators bilingual in Chinese and English (A, B, C) rated two direction-specific packs of $100$ WMT22 sentences each. The public repository includes the written guidelines, row-level CSV form templates, and system-to-slot mapping examples.

\begin{table*}[!t]
\centering
\footnotesize
\setlength{\tabcolsep}{4pt}
\resizebox{\textwidth}{!}{%
\begin{tabular}{l cccc}
\toprule
 & \textbf{A} & \textbf{B} & \textbf{C} & \textbf{Agg.} \\
\midrule
\rowcolor{EVGroup}
\multicolumn{5}{l}{\emph{Panel A: \mbox{En$\to$Zh}} (N=100)} \\
$\Delta$ Adequacy (EV $-$ Ratio), mean & $+0.19$ & $+0.17$ & $+0.19$ & $+0.18$ \\
$\Delta$ Fluency (EV $-$ Ratio), mean  & $+0.18$ & $+0.20$ & $+0.16$ & $+0.18$ \\
Preference EV / Ratio / Tie & 28/19/53 & 28/16/56 & 30/27/43 & 28/19/53 \\
Wilcoxon $p$ (Adequacy) & $0.038^{\ast}$  & $0.070$  & $0.065$  & --- \\
Wilcoxon $p$ (Fluency)  & $0.056$  & $0.035^{\ast}$  & $0.135$  & --- \\
\midrule
\rowcolor{EVGroup}
\multicolumn{5}{l}{\emph{Panel B: \mbox{Zh$\to$En}} (N=100)} \\
$\Delta$ Adequacy (EV $-$ Ratio), mean & $\mathbf{+0.54}$ & $\mathbf{+0.52}$ & $\mathbf{+0.43}$ & $\mathbf{+0.50}$ \\
$\Delta$ Fluency (EV $-$ Ratio), mean  & $+0.12$ & $+0.09$ & $+0.12$ & $+0.11$ \\
Preference EV / Ratio / Tie & 43/27/30 & 38/23/39 & 52/32/16 & 45/25/30 \\
Wilcoxon $p$ (Adequacy) & $0.0001^{\ast\ast\ast}$ & $0.0007^{\ast\ast\ast}$ & $0.003^{\ast\ast}$ & $0.001^{\ast\ast}$ \\
Wilcoxon $p$ (Fluency)  & $0.52$ & $0.79$ & $0.44$ & $0.95$ \\
\midrule
\rowcolor{EVGroup}
\multicolumn{5}{l}{\emph{Agreement and metric correlation:}} \\
Fleiss' $\kappa$ (pref., \mbox{En$\to$Zh} / \mbox{Zh$\to$En}) & \multicolumn{4}{c}{$0.574$ / $0.522$} \\
Kendall's $W$ (score ranks, range) & \multicolumn{4}{c}{\mbox{En$\to$Zh}: $0.68$--$0.74$; \mbox{Zh$\to$En}: $0.81$--$0.90$} \\
Majority-pref.\ sign-test $p$ & \multicolumn{4}{c}{\mbox{En$\to$Zh}: $0.24$; \mbox{Zh$\to$En}: $0.022^{\ast}$} \\
$\rho(\Delta\mathrm{COMET}, \Delta\mathrm{Adeq})$ \mbox{En$\to$Zh} / \mbox{Zh$\to$En} & \multicolumn{4}{c}{$0.637$ / $0.689$} \\
$\rho(\Delta\mathrm{COMET}, \Delta\mathrm{Flu})$ \mbox{En$\to$Zh} / \mbox{Zh$\to$En} & \multicolumn{4}{c}{$0.461$ / $0.455$} \\
\bottomrule
\end{tabular}%
}
\caption{\textbf{Detailed human-evaluation statistics.} The first three columns report annotator-level results. The aggregate preference uses per-sentence majority vote; aggregate \mbox{Zh$\to$En} adequacy uses the paired Wilcoxon test on the three-expert mean.}
\label{tab:human_eval_details}
\end{table*}

\paragraph{Form schema.} Each direction uses one spreadsheet with one row per sentence and eleven columns: \texttt{ID}, \texttt{Source}, \texttt{Reference}, anonymised outputs in \texttt{System A} and \texttt{System B}, four 1--5 rating cells for adequacy and fluency, \texttt{Preference} (A / B / Tie), and free-text \texttt{Notes}. An independent Bernoulli(0.5) flip maps each system to slot A or B for every row; annotators never see the mapping. Each direction contains $25$ EV-best cases by COMET, $25$ Ratio-best cases, $25$ near ties with the smallest nonzero $|\Delta\mathrm{COMET}|$, and $25$ random cases.

\paragraph{Rating dimensions.} \emph{Adequacy} measures how completely and accurately the translation conveys the source meaning, on a 1--5 scale: 5 = all source information conveyed accurately with no omissions, additions, or errors; 4 = nearly all information, with only minor omissions or imprecision; 3 = main idea preserved but visible information loss; 2 = only some information correct, with major mistranslations or omissions; 1 = unrelated or unintelligible. \emph{Fluency} measures how natural and grammatical the target text is, on a 1--5 scale: 5 = reads as native; 4 = mostly natural with occasional unnaturalness; 3 = understandable but with multiple unnatural expressions; 2 = frequent grammatical errors, hard to parse; 1 = severe grammar errors or incoherent text. Adequacy and fluency are scored independently.

\paragraph{Preference rule.} \texttt{Preference} is a single A / B / Tie choice that combines both dimensions. If one system dominates on both adequacy and fluency, pick that system; if the two trade off, pick the more competent overall translation; if the difference is tiny (e.g.\ $4/4$ vs $4/5$), pick \texttt{Tie}.

\paragraph{Blinding and independence.} Per-row slot randomisation prevents annotators from inferring system identity from row order. The three annotators worked independently on the same $100$ sentences per direction with no cross-annotator discussion during rating. No annotator was shown the other direction's ratings before completing their own. The study used public WMT22 sentences and did not collect sensitive personal data. Annotators consented to the use of their anonymised ratings and aggregate statistics for research reporting.

\paragraph{Special-case handling.} Empty outputs (no translation produced) are scored $1/1$ on both adequacy and fluency. Outputs with repeated fragments (the same phrase repeated) receive a fluency penalty. Unusual observations can be recorded in the free-text \texttt{Notes} column and do not affect the numeric ratings.

\paragraph{Compensation and pacing.} Annotators were compensated at local professional-translator hourly rates. Annotation averaged 1--2 minutes per sentence; each 100-sentence batch was completed in 2--3 sittings, with the directions rated on separate days.

\end{document}